\pdfoutput=1

\PassOptionsToPackage{table, prologue}{xcolor}
\PassOptionsToPackage{backref=page}{hyperref}

\documentclass[11pt]{article}

\usepackage{EMNLP2022}

\usepackage{times}
\usepackage{latexsym}
\usepackage{todonotes}
\usepackage{enumitem}
\usepackage{lipsum}
\usepackage{booktabs}
\usepackage{array}
\usepackage{multirow}
\usepackage{graphicx}
\usepackage[normalem]{ulem}

\usepackage{etoolbox}
\usepackage{todonotes}
\usepackage{amsmath}
\usepackage{color}
\usepackage{tcolorbox}
\usepackage{CJK}
\usepackage{adjustbox}
\usepackage{caption}
\usepackage{subcaption}
\usepackage{bbm}
\usepackage{verbatim}

\usepackage[T1]{fontenc}

\usepackage[utf8]{inputenc}

\usepackage{microtype}

\usepackage{inconsolata}

\usepackage{xspace}

\newcommand\DATASET{\textsc{Spiced}}
\newcommand\ds{\DATASET\xspace}

\newcommand\DATASETFULL{\textsc{Scientific Paraphrase and Information ChangE Dataset}}

\newcommand\SCORE{\textsc{IMS}}
\newcommand\SCOREFULL{\textsc{Information Matching Score}}

\newcommand{\Sref}[1]{\S\ref{#1}}

\newcommand{\cameraready}[1]{}

\title{Modeling Information Change in Science Communication with Semantically Matched Paraphrases}
\author{Dustin Wright\textsuperscript{$\flat$}\thanks{~~denotes equal contribution}\hspace{0.5cm} Jiaxin Pei\textsuperscript{$\sharp$}\footnotemark[1]\hspace{0.5cm} David Jurgens\textsuperscript{$\sharp$}\hspace{0.5cm} Isabelle Augenstein\textsuperscript{$\flat$}\hspace{0.5cm} \\
  \textsuperscript{$\flat$}Dept. of Computer Science, University of Copenhagen, Denmark \\
  \textsuperscript{$\sharp$}School of Information, University of Michigan, Ann Arbor, MI, USA \\
  \texttt{\{dw,augenstein\}@di.ku.dk}\\
  \texttt{\{pedropei,jurgens\}@umich.edu}}

\begin{document}
\maketitle
\begin{abstract}
Whether the media faithfully communicate scientific information has long been a core issue to the science community. Automatically identifying paraphrased scientific findings could enable large-scale tracking and analysis of information changes in the science communication process, but this requires systems to understand the similarity between scientific information across multiple domains. To this end, we present the \DATASETFULL{} (\DATASET{}), the first paraphrase dataset of scientific findings annotated for degree of information change. \DATASET{} contains 6,000 scientific finding pairs extracted from news stories, social media discussions, and full texts of original papers. We demonstrate that \DATASET{} poses a challenging task and that models trained on \DATASET{} improve downstream performance on evidence retrieval for fact checking of real-world scientific claims.  Finally, we show that models trained on \DATASET{} can reveal large-scale trends in the degrees to which people and organizations faithfully communicate new scientific findings.  Data, code, and pre-trained models are available at \url{http://www.copenlu.com/publication/2022_emnlp_wright/}.
\end{abstract}

\section{Introduction}

\begin{figure}[t]
  \centering
    \includegraphics[width=\linewidth]{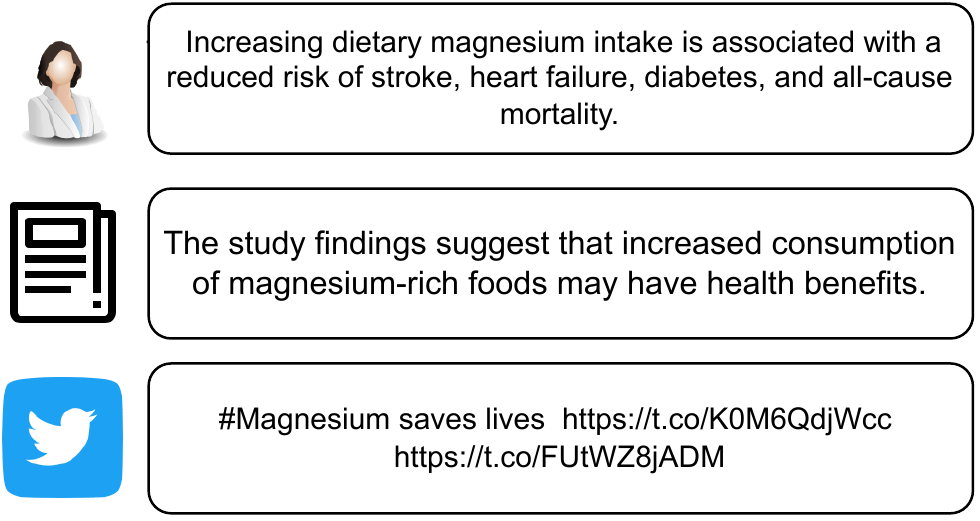}
    \caption{We are interested in measuring the information similarity of statements about scientific findings between different sources, including scientific papers, news, and tweets, shown here with real examples. The finding in this figure comes from~\citet{fang2016dietary} and the news quote is from \href{https://www.reuters.com/article/us-health-diet-magnesium-idUSKBN14J1DG}{this Reuters story}.} 
    \label{fig:claim_generation}
\end{figure}

Science communication disseminates scholarly information to audiences outside the research community, such as the public and policymakers \citep{national2017communicating}. This process usually involves translating highly technical language to non-technical, less-formal language that is engaging and easily understandable for lay people \citep{Salita2015WritingFL}. The public relies on the media to learn about new scientific findings, and media portrayals of science affect people's trust in science while at the same time influencing their future actions \citep{gustafson2019effects,fischhoff2012communicating,kuru2021effects}. However, not all scientific communication accurately conveys the original information, as shown in \autoref{fig:claim_generation}. Identifying cases where scientific information has changed is a critical but challenging task due to the complex translating and paraphrasing done by effective communicators. Our work introduces a new task of measuring scientific information change, and through developing new data and models aims to address the gap in studying faithful scientific communication.



Though efforts exist to track and flag when popular media misrepresent science,\footnote{See e.g. \url{https://www.healthnewsreview.org/} and \url{https://sciencefeedback.co/}} the sheer volume of new studies, reporting, and online engagement make purely manual efforts both intractable and unattractive. Existing studies in NLP to help automate the study of science communication have examined exaggeration \citep{wright2021semi}, certainty~\cite{DBLP:conf/emnlp/PeiJ21}, and fact checking~\cite{DBLP:conf/bionlp/BoissonnetSPV22,DBLP:conf/acl/0001WLKCAW22}, among others. However, these studies skip over the key first step needed to compare scientific texts for information change: automatically identifying content from both sources which describe the \textbf{same} scientific finding. In other words, to answer relevant questions about and analyze changes in scientific information at scale, one must first be able to point to which original information is being communicated in a new way. 



To enable automated analysis of science communication, this work offers the following \textbf{contributions} (marked by \textbf{C}). First, we present the \DATASETFULL{} dataset (\DATASET{}), a manually annotated dataset of paired scientific findings from news articles, tweets, and scientific papers (\textbf{C1}, \S\ref{sec:datset}). 
\DATASET{} has the following merits: (1) existing datasets focus purely on semantic similarity, while \DATASET{} focuses on differences in the \textit{information} communicated in scientific findings; (2) scientific text datasets tend to focus solely on titles or paper abstracts,  while \DATASET{} includes sentences extracted from the full-text of papers and news articles; (3) \DATASET{} is largely multi-domain, covering the 4 broad scientific fields that get the most media attention (namely: medicine, biology, computer science, and psychology) and includes data from the whole science communication pipeline, from research articles to science news and social media discussions. 

In addition to extensively benchmarking the performance of current models on \DATASET{} (\textbf{C2}, \S\ref{sec:benchmarking}), we demonstrate that the dataset enables multiple downstream applications. In particular, we demonstrate how models trained on \DATASET{} improve zero-shot performance on the task of sentence-level evidence retrieval for verifying real-world claims about scientific topics (\textbf{C3}, \S\ref{sec:evidence-retrieval}), and perform an applied analysis on unlabelled tweets and news articles where we show (1) media tend to exaggerate findings in the limitations sections of papers; (2) press releases and SciTech tend to have less informational change than general news outlets; and (3) organizations' Twitter accounts tend to discuss science more faithfully than verified users on Twitter and users with more followers (\textbf{C4}, \S\ref{sec:social-media-analysis}).

\section{Related Work}
The analysis of scientific communication directly relates to fact checking, scientific language analysis, and semantic textual similarity. We briefly highlight our connections to these.

\paragraph{Fact Checking} Automatic fact checking is concerned with verifying whether or not a given claim is true, and has been studied extensively in multiple domains~\cite{DBLP:conf/naacl/ThorneVCM18,DBLP:conf/emnlp/AugensteinLWLHH19} including science~\cite{DBLP:conf/emnlp/WaddenLLWZCH20,DBLP:conf/bionlp/BoissonnetSPV22,DBLP:conf/acl/0001WLKCAW22}. Fact checking focuses on a specific type of information change, namely veracity. Additionally, the task generally assumes access to pre-existing knowledge resources, such as Wikipedia or PubMed, from which evidence can be retrieved that either supports or refutes a given claim. Our task is concerned with a more general type of information change beyond categorical falsehood and is a required task to complete prior to performing any kind of fact check.

\paragraph{Scientific Language Analysis}
Automating tasks beneficial for understanding changes in scientific information between the published literature and media is a growing area of research~\cite{wright2021semi,DBLP:conf/emnlp/PeiJ21,DBLP:conf/bionlp/BoissonnetSPV22,DBLP:conf/icwsm/DaiSW20,august2020writing,DBLP:conf/acl/TanL14,DBLP:conf/emnlp/VadapalliSPSV18,DBLP:conf/chi/AugustCHSR20,DBLP:conf/lrec/GinevM20}. The three tasks most related to our work are 
understanding writing strategies for science communication~\cite{august2020writing}, detecting changes in certainty~\cite{DBLP:conf/emnlp/PeiJ21}, and detecting changes in causal claim strength i.e. exaggeration~\cite{wright2021semi}. However, studying these requires access to paired scientific findings. To be able to do so at scale will require the ability to pair such findings automatically.


\paragraph{Semantic Similarity} The topic of semantic similarity is well-studied in NLP. Several datasets exist with explicit similarity labels, many of which come from SemEval STS shared tasks~\citep[e.g.,][]{DBLP:journals/corr/abs-1708-00055} and paraphrasing datasets~\cite{DBLP:conf/naacl/GanitkevitchDC13}. It is possible to build unlabelled datasets of semantic similarity automatically, which is the main method that has been used for scientific texts~\cite{DBLP:conf/acl/CohanFBDW20,DBLP:conf/acl/LoWNKW20}. 
However, such datasets fail to capture more subtle aspects of similarity, particularly when the focus is solely on the scientific findings conveyed by a sentence (see \autoref{sec:info-change-supplement}). 
And as we will show, approaches based on these datasets are insufficient for the task we are concerned with in this work, motivating the need for a new resource.

\section{\DATASET}
\label{sec:datset}

We introduce \ds, a new large-scale dataset of \textit{scientific findings} paired with how they are communicated in news and social media. Communicating scientific findings is known to have a broad impact on public attitudes \cite{weigold2001communicating} and to influence behavior, e.g., the way vaccines are framed in the media has an effect on vaccine uptake \cite{kuru2021effects}. Building upon prior work in NLP \cite{wright2021citeworth,DBLP:conf/emnlp/PeiJ21,sumner2014association,bratton2019association}, we define a scientific finding as \textbf{a statement that describes a particular research output of a scientific study, which could be a result, conclusion, product, etc.} 
This general definition holds across fields; for example, many findings from medicine and psychology report on effects on some dependent variable via manipulation of an independent variable, while in computer science many findings are related to new systems, algorithms, or methods. 
Following, we describe how the pairs of scientific findings were selected and annotated.



\subsection{Data Collection}

An initial dataset of unlabelled pairs of scientific communications was collected through Altmetric  ({\small{\url{https://www.altmetric.com/}}}) a platform tracking mentions of scientific articles online. 
This initial pool contains 17,668 scientific papers, 41,388 paired news articles, and 733,755 tweets---note that a single paper may be communicated about multiple times. The scientific findings were extracted in different ways for each source. Similar to \citet{Prabhakaran2016PredictingTR}, we fine-tune a RoBERTa~\cite{DBLP:journals/corr/abs-1907-11692} model to classify sentences into methods, background, objective, results and conclusions using 200K paper abstracts from PubMed that had been self-labeled with these categories~\cite{canese2013pubmed}. This sentence classifier attained 0.92 F1 score on a held-out 10\% sample (details in \autoref{sec:parser-supplement}) and then the classifier was applied to each sentence of the news stories and paper fulltexts. 
Given the domain difference between scientific abstracts and news, we additionally manually annotated a sample of 100 extracted conclusions; we find that the precision of the classifier is 0.88, suggesting that it is able to accurately identify scientific findings in news as well. We extract each sentence classified as ``result'' or ``conclusion'' and create pairs with each finding sentence from news articles written about it. This yields 45.7M potential pairs of $\langle$news, paper$\rangle$ findings. For tweets, we take full tweets as is, yielding 35.6M potential pairs  of $\langle$tweet, paper$\rangle$ findings.


\subsection{Data sampling}
\label{sec:news-pairs}
\begin{table*}[t]
\small

\newcommand{\tabincell}[2]{\begin{tabular}{@{}#1@{}}#2\end{tabular}}
\resizebox{0.99\textwidth}{!}{
\begin{tabular}{p{62mm}p{62mm}cc}
\textbf{Paper finding } & \textbf{News Finding} & \textbf{Similarity Score} & \textbf{\SCORE{}} \\
\midrule
However, the consistency of the erythritol results in both the central adiposity and usual glycemia comparisons lends strength to the findings, and the cluster of metabolites has biological plausibility. & Young adults who exhibited central adiposity gain over the course of 35 weeks had plasma erythritol levels 15-times higher at baseline than those with stable adiposity over the same period. & 0.88 & 1 \\ \midrule
Our results showed that most of the official adult-onset men began their antisocial activities during early childhood. & Beckley, who is in the department of psychology and neuroscience at Duke, said the adult-onset group had a history of anti-social behavior back to childhood, but reported committing relatively fewer crimes. & 0.38 & 4.4 \\
\end{tabular}
}
\caption{ Annotated information matching score (IMS) and the similarity score estimated by SBERT~\cite{reimers-2019-sentence-bert} for selected finding pairs from \DATASET{}. These examples demonstrate that simple similarity scores may not reflect whether the two sentences are covering the same scientific finding.}
\label{tab:example_scores}
\end{table*}

  

Pairing every finding from a news story with every finding from its matched paper results in an untenable amount of data to annotate. Additionally, it has been shown that proper data selection can reduce the need to annotate every possible sample~\cite{DBLP:journals/neco/MacKay92b,DBLP:conf/cvpr/HolubPB08,DBLP:journals/corr/abs-1112-5745}. Therefore, to obtain a sample of paired findings covering a range of similarities, we first filter our pool of unlabelled matched findings based on the semantics 
using Sentence-BERT (SBERT, \citet{reimers-2019-sentence-bert}), a Siamese BERT network trained for semantic text similarity, trained on over 1B sentence pairs (see \autoref{sec:full-model-descriptions} for further details). We use this model to score pairs of findings from news articles and papers based on their embeddings' cosine similarity and conduct a pilot study to determine which data to annotate. 

  

For the pilot, we sample 400 pairs evenly for every $0.05$ increment bucket in the range $[0,1]$ of similarity scores (20 per bucket). 
Each sample is annotated by two of the authors of this study with a binary label of ``matching'' vs ``not matching'', yielding a Krippendorff's alpha of $0.73$.\footnote{Note that many discussions about what constitutes matching vs. not matching were had in pilot work, leading to high agreement.} From this sample, we observed that there were no matches below 0.3 and only 2 ambiguous matches below 0.4. At the same time, the vast majority of samples from the entire dataset have a similarity score of less than 0.4. Additionally, above 0.9 we saw that each pair was essentially equivalent. Given the distribution of matched findings across the similarity scale, in order to balance the number of annotations we can acquire, the yield of positive samples, and the sample difficulty, we sampled data as follows based on their cosine similarity:

\begin{itemize}[noitemsep,nolistsep]
    \item Below $0.4$ = automatically unmatched.
    \item Above $0.9$ with a Jaccard index above $0.5$ = automatically matched.
    \item Sample an equal number of pairs from each $0.05$ increment bin between $0.4$ and $0.9$ for human expert annotation.
\end{itemize}

We sample 600 $\langle$news, paper$\rangle$ finding pairs from the four fields which receive the most media attention (medicine, biology, computer science, and psychology) using this method. 
This yields 2,400 pairs to be annotated. For extensive details on the pilot annotation and visualizations, see \autoref{sec:pilot-annotation-details}. 

We follow a similar procedure to sample pairs from papers and Twitter for annotation. However, rather than use the SBERT similarity scores, we instead first obtain annotations for news pairs using the scheme to be described later in \S\ref{sec:annotation-scheme} in order to train an initial model on our task (CiteBERT, \citealt{wright2021citeworth}). We then use the trained model to obtain scores in the range [0,1] for each pair and sample an equal number of pairs from bins in 0.05 increments, for a total of 1,200 pairs (300 from each field of interest). 

\subsection{Finding Matching Annotation}
\label{sec:annotation-scheme}
We perform our final annotation based on the sampling scheme above using the Prolific platform ({\small{\url{https://www.prolific.co/}}}) as it allows prescreening annotators by educational background.
We require each annotator to have at least a bachelor's degree in a relevant field to work on the task. 
Annotators are asked to label ``whether the two sentences are discussing the same scientific finding'' for 50 finding pairs with a 5-point Likert schema where each value indicates that 
``The information in the findings is...''
(1): Completely different
(2): Mostly different
(3): Somewhat similar
(4): Mostly the same, or
(5): Completely the same.
%
%
%
%
See  \autoref{sec:experimented-annotations-supplement} for details  of how this rating scale was decided.
We call this the \SCOREFULL{} (\SCORE{}) of a pair of findings. Annotation was performed using \textsc{Potato} \cite{pei2022potato}. Full annotation instructions and details are listed in \autoref{sec:annotation-instructions}. Notably, annotators were instructed to mark how similar the information in the \textit{findings} was, as opposed to how similar the sentences are. Further, they were instructed to ignore extraneous information like ``The scientists show...'' and ``our experiments demonstrate...''. 




\paragraph{Post processing} 
To improve the reliability of the annotations, we use MACE~\cite{DBLP:conf/naacl/HovyBVH13} to estimate the competence score of each annotator and removed the labels from the annotators with the lowest competence scores. 
\cameraready{MACE is a Bayesian method that learns distributions over true class labels and annotator competence based on crowd-sourced labels. }
We further manually examine pairs with the most diverse labels (standard deviation of ratings $>$1.2) and manually replace the outliers with our expert annotations. The overall Krippendoff's $\alpha$ is 0.52, 0.57, 0.53, and 0.52 for CS, Medicine, Biology, and Psychology respectively, indicating that the final labels are reliable. While many annotators considered the task challenging, our quality control strategies allow us to collect reliable annotations.\footnote{For example, one participant commented ``It was pretty hard to consider both the statements and their context then comparing them for similarities, but i enjoyed it''} For all the annotated pairs, we average the ratings as the final similarity score. In addition to the 3,600 manually annotated pairs, we include an extra 2,400 automatically annotated pairs as determined in \S\ref{sec:news-pairs} (unmatched pairs get an IMS of 1, matched pairs get an IMS of 5), for a total of 6,000 pairs. 
Given that there can be multiple pairs from a single news-paper pair, to avoid overlaps between training and test sets, we split the dataset 80\%/10\%/10\% based on the paper DOI and balance across subjects. Further dataset details in \autoref{sec:dataset-details-supplement}

\paragraph{Selected Examples}
To highlight the difficulty of \DATASET{}, we show a pair of samples from our final dataset in  \autoref{tab:example_scores}. 
The \SCORE{} is compared to the cosine similarity between embeddings produced by SBERT. For the first case, SBERT presumably picks up on similarities in the discussed topics, such as erythritol and its relationship to adiposity, but the paper finding is concerned with the consistency of results and its biological implications while the news finding explicitly mentions a relationship between erythritol and adiposity. The second case expresses the opposite effect; the news finding contains a lot of extraneous information for context, but one of the core findings it expresses is the same as the paper finding, giving it a high rating in \DATASET{}.



\paragraph{Comparison with existing datasets} 
To further characterize the difficulty of \DATASET{} compared to existing datasets, we show the average normalized edit distance between matching pairs in \DATASET{}, STSB~\cite{DBLP:journals/corr/abs-1708-00055}, and SNLI~\cite{snli:emnlp2015} (see \autoref{sec:metrics-supplement} for the calculation). STSB is a semantic text similarity dataset consisting of pairs of sentences scored with their semantic similarity, sourced from multiple SemEval shared tasks. SNLI is a natural language inference corpus, and consists of pairs of sentences labeled for if they entail each other, contradict each other, or are neutral.
We calculated the mean normalized edit distance across all pairs of \textit{matching} sentences in each dataset's training data;  For \DATASET{} and STSB, pairs are considered matching if their IMS or similarity score is greater than 3, respectively. For SNLI, pairs are considered matching if the label is ``entailment''.

\begin{table}
    \def\arraystretch{1.2}
    \centering
    \fontsize{10}{10}\selectfont
    \begin{tabular}{c c c c c}
    \toprule 
    STSB & SNLI & \DATASET{} & News & Tweets\\
    \midrule 

         $0.401$ & $0.631$ & $\mathbf{0.726}$ & $\mathit{0.712}$ & $\mathit{0.749}$\\
    
    \bottomrule 

    \end{tabular}
    \caption{The average normalized edit distance between matching pairs for various datasets shows that \DATASET{} includes more pairs that are lexically dissimilar. For \DATASET{} and STSB, pairs are considered matching if their similarity score is greater than 3. For SNLI, pairs are considered matching if the label is ``entailment''.}
    \label{tab:edit-distance}
\end{table}

We find that there is a much greater lexical difference between the matching pairs in \DATASET{} (0.726) than existing general domain paired text datasets (0.401 for STSB and 0.631 for SNLI). This gap between STSB and \ds also emphasizes the difference between traditional semantic textual similarity tasks and the information change task we describe here. Within \ds, Twitter pairs had a higher distance (0.749) than news pairs (0.712), suggesting stronger domain differences. For qualitative examples showing the difference between \DATASET{} and STSB, see \autoref{sec:info-change-supplement}.

\paragraph{Relationship of \DATASET{} to Fact Checking}
The task introduced by \DATASET{} 
captures information change more broadly than veracity as in automatic fact checking, as the task is concerned with the degree to which two sentences describe the same scientific information---indeed, two similar sentences may describe the same information equally poorly. Our task is similar to the sentence selection stage in the fact checking pipeline, and we later demonstrate that models trained on \DATASET{} data are useful for this task for science in \autoref{sec:evidence-retrieval}. However, our task and annotation are agnostic to whether a pair of sentences entail one another. This is especially useful if one wants to compare how a particular finding is presented across different media. Fact-checking datasets are also explicitly constructed to contain claims which are about a single piece of information---\DATASET{} is not restricted in this way, focusing on a more general type of information change beyond categorical falsehood. Finally, we note two more unique features of \DATASET{}: 1) \DATASET{} contains naturally occurring sentences, while fact checking datasets like FEVER and SciFact often contain manually written claims. 2) The combination of domains in \DATASET{} is unique; sentences are paired between (news, science) and (tweets, science), and these pairings don’t exist currently.

  

\section{Scientific Information Change Models}
\label{sec:benchmarking}

\begin{figure*}[t]
         \centering
     \begin{subfigure}[b]{0.495\textwidth}
         \centering
         \includegraphics[width=\textwidth]{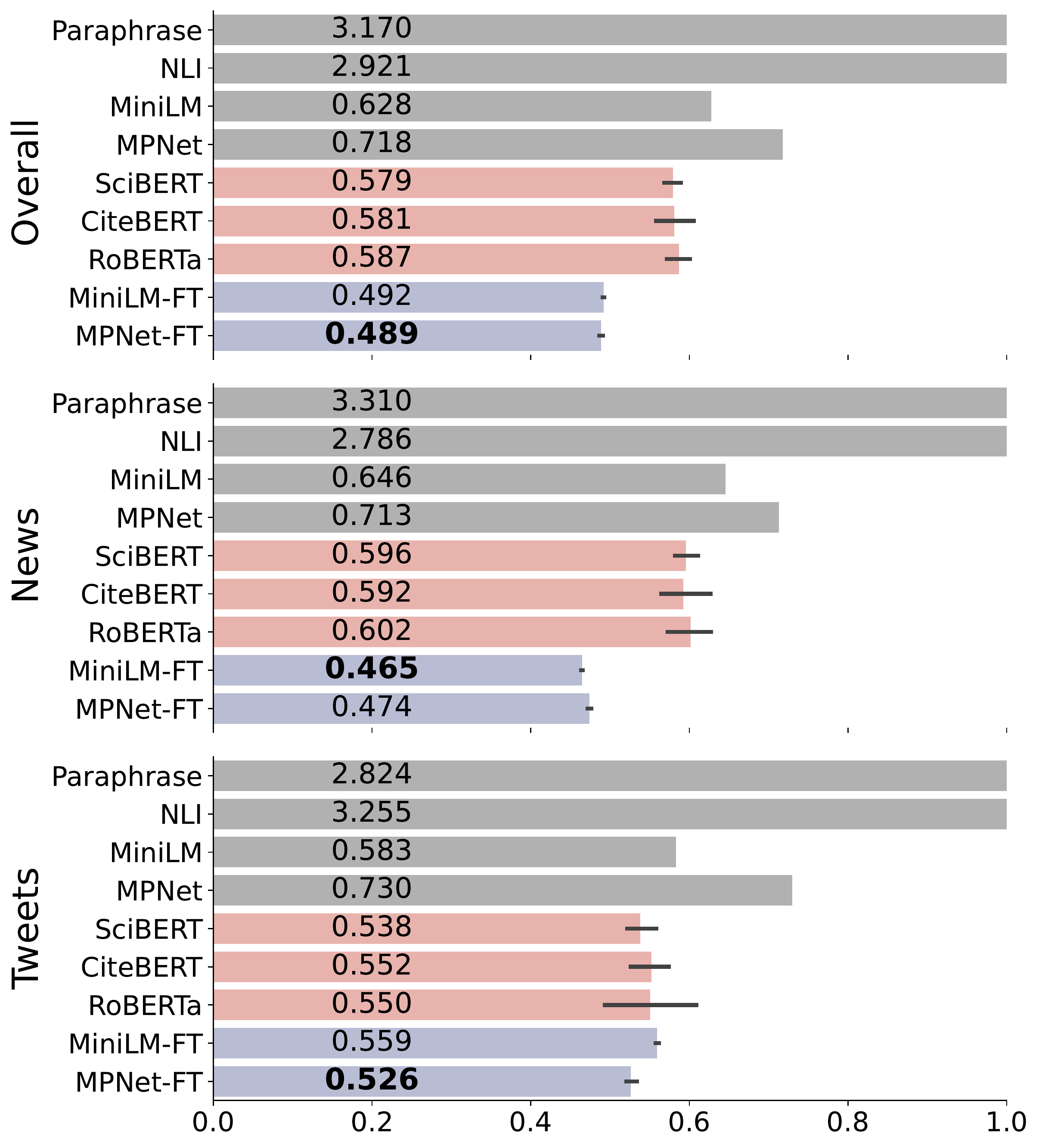}
         \caption{Mean Squared Error}
         \label{fig:mse-baselines}
     \end{subfigure}
     \hfill
     \begin{subfigure}[b]{0.495\textwidth}
         \centering
         \includegraphics[width=\textwidth]{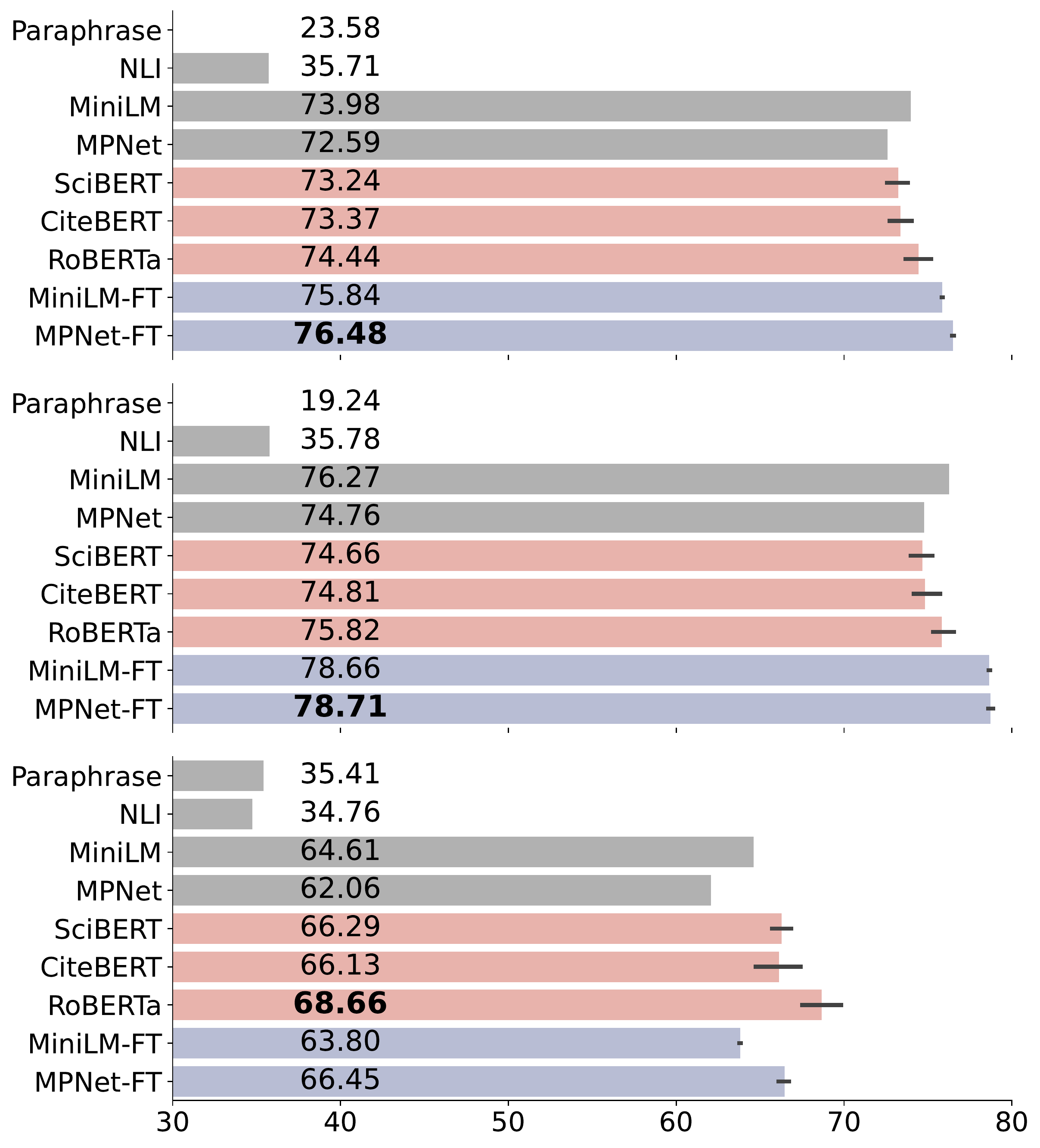}
         \caption{Pearson correlation ($r$)}
         \label{fig:r-baselines}
     \end{subfigure}
 
    \caption{(a) Mean Squared Error (MSE, $\downarrow$ better) and (b) Pearson correlation ($r$, $\uparrow$ better) on the test set of \DATASET{}. Grey = zero-shot transfer models, red = MLM models fine-tuned on \DATASET{},blue = SBERT models fine-tuned on \DATASET{}. Results are averaged across 5 random seeds. Best results are given in bold.}
    \label{fig:baseline-performance}
\end{figure*}

We now use \ds to evaluate models for estimating the IMS of finding pairs in two settings: zero-shot transfer and supervised fine-tuning. 

\subsection{Experimental setup} 

We use the following four models to estimate zero-shot transfer performance.
\textbf{Paraphrase}: RoBERTa~\cite{DBLP:journals/corr/abs-1907-11692} pre-trained for paraphrase detection on an adversarial paraphrasing task~\cite{nighojkar-licato-2021-improving}. We convert the output probability of a pair being a paraphrase to the range [1,5] for comparison with our labels. \textbf{Natural Language Inference (NLI)}: RoBERTa pre-trained on a wide range of NLI datasets~\cite{nie-etal-2020-adversarial}. The final score is the model's measured probability of entailment mapped to the range [1,5]. \textbf{MiniLM}: SBERT with MiniLM as the base network~\cite{DBLP:conf/nips/WangW0B0020}; we obtain sentence embeddings for pairs of findings and measure the cosine similarity between these two embeddings, clip the lowest score to 0, and convert this score to the range [1,5]. Note that this model was trained on over 1B sentence pairs, including from scientific text, using a contrastive learning approach where the embeddings of sentences known to be similar are trained to be closer than the embeddings of negatively sampled sentences. SBERT models represent a very strong baseline on this task, and have been used in the context of other matching tasks for fact checking including detecting previously fact-checked claims~\cite{shaar-etal-2020-known}. \textbf{MPNet}: The same setting and training data as MiniLM but with MPNet as the base network~\cite{DBLP:conf/nips/Song0QLL20}. 

We fine-tune the following six models on \ds to estimate IMS as a comparison with zero-shot transfer.
\begin{itemize}[noitemsep,nolistsep]
\item{\textbf{MiniLM-FT}: The same MiniLM model from the zero-shot transfer setup but further fine-tuned on \DATASET{}. The training objective is to minimize the distance between the IMS and the cosine similarity of the output embeddings of the pair of findings.}
\item{\textbf{MPNet-FT}: The same setup as MiniLM-FT but using MPNet as the base network.}
\item{\textbf{RoBERTa}: The RoBERTa~\cite{DBLP:journals/corr/abs-1907-11692} base model; We perform a regression task where the model is trained to minimize the mean-squared error between the prediction and IMS.}
\item{\textbf{SciBERT}: A transformer model trained using masked language modeling on a large corpus of scientific text~\cite{beltagy-etal-2019-scibert}. The fine-tuning setup is the same as for the RoBERTa model.}
\item{\textbf{CiteBERT}: A SciBERT model further fine-tuned on the task of citation detection, and was shown to have improved performance on downstream tasks using scientific text~\cite{wright2021citeworth}. The training setup is the same as for the RoBERTa model.}
\end{itemize}

Please see \autoref{sec:full-model-descriptions} for further details on the models and pretraining methods.
For the fine-tuned models, we train on the entire training set of \DATASET{}, including both news findings and tweets\cameraready{ and show test performance overall and split between news and tweets}. For the test set we only use manually annotated pairs. Performance is measured in terms of mean-squared error (MSE) and Pearson correlation ($r$) (definitions of all metrics in \autoref{sec:metrics-supplement}). All results are reported as the average and standard deviation for each model across 5 random seeds. 

\subsection{Results}

Paraphrase detection and natural language inference models perform very poorly for zero-shot transfer on this task (\autoref{fig:baseline-performance}, grey bars), with NLI having slightly better transfer, supporting our hypothesis that transferring from existing tasks to this domain is challenging. Fine-tuned models with Masked Language Model (MLM) pretraining can learn the task decently well (\autoref{fig:baseline-performance}, red bars), but surprisingly RoBERTa performs just as well as SciBERT and CiteBERT which were specifically pretrained on scientific texts. We posit that this could be due to the fact that RoBERTa was pretrained on a wider range of texts that are reflective of the domains in \DATASET{}, including news texts, while SciBERT and CiteBERT were trained solely on scientific papers. 

SBERT models trained on large amounts of pretraining sentences perform well in the zero-shot transfer setup, with the MiniLM based model outperforming MPNet. The best setup was using SBERT fine-tuned on \DATASET{} (\autoref{fig:baseline-performance}, blue bars), which yields up to 3.9 points gained overall in Pearson correlation and a reduction of 0.3 in terms of MSE (MPNet to MPNet-FT). We also note that there is a large gap between performance on this data and general semantic similarity datasets such as STSB, which see correlation scores in the 90s. As such, there is potentially much room to grow in terms of raw performance on this dataset.

Models performed worse for pairs with tweets versus those from news (Appendix \autoref{tab:baseline-findings-matching}). This performance difference is in line with our expectations, as there is a large domain shift between tweets and scientific texts and our base models were not exposed to tweets during pre-training. All models, including the zero-shot transfer SBERT models, perform much worse on that split of the data. Additionally, we only see minor gains in performance in terms of MSE for MiniLM when fine-tuned on tweets. We see larger gains for MPNet. Interestingly, the best performance (Pearson $r$) for Tweets is RoBERTa, though the overall MSE is still best for MPNet-FT. We show extended benchmarking in \autoref{sec:extended-benchmarking} and the top-5 errors for RoBERTa and MPNet-FT in \autoref{sec:error-examples}.

\section{Application: Zero-Shot Evidence Retrieval for Scientific Fact Checking}
\label{sec:evidence-retrieval}

Accurately measuring the similarity of scientific findings written in different domains enables a wide range of downstream analyses and tasks. 
As a first task, we consider evidence retrieval for scientific fact checking of real-world scientific claims. In general, automatic fact checking consists of retrieving relevant evidence for a given claim and predicting if that evidence supports or refutes the claim. 
We test the ability of models trained on \DATASET{} to perform the evidence retrieval task in a zero-shot setting. In this, we use the models as is, with no further fine-tuning on any evidence retrieval data. We consider two fact checking datasets: CoVERT~\cite{DBLP:journals/corr/abs-2204-12164} is a dataset of scientific claims sourced from Twitter, mostly in the domain of biomedicine. 
We use the 300 claims and the 717 unique evidence sentences in the corpus in our experiment. COVID-Fact~\cite{DBLP:conf/acl/SaakyanCM20} is a semi-automatically curated dataset of claims related to COVID-19 sourced from Reddit. 
The corpus contains 4,086 claims with 3,219 unique evidence sentences. 

\paragraph{Setup}
We compare different models' ability to rank the evidence sentences such that the ground truth evidence for a given claim is ranked highest. We use four models in a zero-shot setting for comparison (MiniLM, MiniLM-FT, MPNet, and MPNet-FT; '-FT' indicates fine-tuning on \DATASET{}), and show results with the unsupervised BM25 \cite{Robertson1994OkapiAT}, a widely used bag-of-words retrieval model. 
We report retrieval results in terms of mean average precision (MAP) and mean reciprocal rank (MRR), and average the results for models fine-tuned on \DATASET{} across 5 random seeds.

\begin{table}[t]
    \setlength{\tabcolsep}{1.5pt}
    \def\arraystretch{1.2}
    \centering
    \rowcolors{2}{gray!10}{white}
    \fontsize{10}{10}\selectfont
    \begin{tabular}{l c c | c c}
    \toprule 
     & \multicolumn{2}{c}{CoVERT} & \multicolumn{2}{c}{COVID-Fact} \\
     \cline{2-5}
     \rowcolor{white}
    Method & MAP & MRR & MAP & MRR \\
    \hline 
BM25 & $12.45_{0.00}$&$20.78_{0.00}$&$35.18_{0.00}$&$52.98_{0.00}$ \\
MiniLM & $26.84_{0.00}$&$37.98_{0.00}$&$50.11_{0.00}$&$64.78_{0.00}$ \\
~~~+ FT & $\mathbf{28.23_{0.08}}$&$\mathbf{40.81_{0.16}}$&$52.66_{0.10}$&$66.91_{0.09}$ \\
MPNet & $25.21_{0.00}$&$35.54_{0.00}$&$52.39_{0.00}$&$66.21_{0.00}$ \\
~~~+ FT & $26.84_{0.19}$&$37.65_{0.32}$&$\mathbf{53.61_{0.33}}$&$\mathbf{67.46_{0.28}}$ \\

    \bottomrule 

    \end{tabular}
    \caption{Mean average precision (MAP) and mean reciprocal rank (MRR) for retrieval on the CoVERT and COVID-Fact datasets. All models are zero-shot i.e. without fine-tuning on the retrieval dataset.} 
    \label{tab:retrieval}
\end{table}

\paragraph{Results}
We find that fine-tuning on \DATASET{} provides consistent gains in retrieval performance on both datasets for both SBERT models (\autoref{tab:retrieval}). 
This performance increase is encouraging, as there are two notable differences between \DATASET{} and the two datasets in our experiment. The first is that the tasks are different: \DATASET{} provides a general scientific information similarity task which proves to be useful for evidence sentence ranking.
The second is that the domains are different: \DATASET{} contains $\langle$news, paper$\rangle$ and $\langle$tweet, paper$\rangle$ pairs, while CoVERT and COVID-Fact have claims from Twitter and Reddit, respectively, paired with evidence in news. Our results show that training on \DATASET{} improves the IR performance of the SBERT models, despite the domain and topic differences from our setting.  

\section{Application: Modeling Information Change in Science Communication}
\label{sec:social-media-analysis}

Whether the media faithfully communicate scientific information has long been a core question to the science community \citep{national2017communicating}. Our dataset and models allow us to conduct a large-scale analysis to study information change in science communication. Here, we focus on three research questions: 
\begin{itemize}[noitemsep]
\item{\textbf{RQ1:} Do findings reported by different types of outlets express different degrees of information change from their respective papers?}
\item{\textbf{RQ2:} Do different types of social media users systematically vary in information change when discussing scientific findings?  }
\item{\textbf{RQ3:}  Which parts of a paper are more likely to be miscommunicated by the media?}
\end{itemize}

RQ1-2 focus on the holistic information change captured in \SCORE{}, while RQ3 focuses on what types of information might be changing.


\begin{figure}[t]
  \centering
    \includegraphics[width=0.45\textwidth]{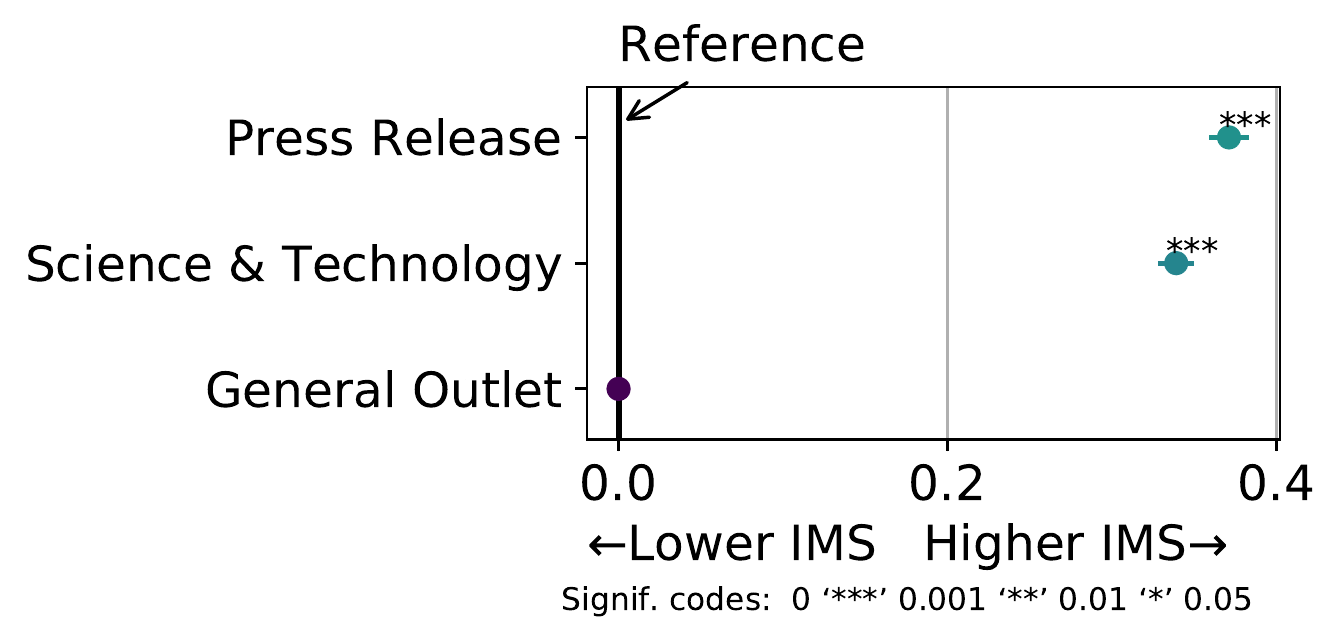}
    \caption{Scientific findings covered by Press Release and SciTech generally have less informational changes compared with findings presented in General Outlets
    }
    \label{fig:news_outlet_type_matching_score}
\end{figure}

\subsection{RQ1: Comparing Media Outlets}
Different types of media target different audiences and tend to report the same issue differently \cite{richardson1990writing, mencher1997news}. While good science journalism requires outlets to prioritize quality, in real practices, journalists may adopt different writing strategies for different types of audiences \citep{roland2009quality}. 
Thus, we investigate if findings reported by different types of outlets express different levels of information change, focusing on three types of outlets: General News (e.g., NYTimes), Press Releases (e.g., Science Daily), and Science \& Technology (e.g., Popular Mechanics).  
We use our best-performing MPNet-FT model to estimate the IMS of over 1B pairs and keep those with IMS $>$ 3, which finally leads to 
1.1M paired findings from 26,784 news stories and 12,147 papers. We then build a linear mixed effect regression model \citep{galecki2013linear} to predict \SCORE{} for matching pairs from news stories and research articles. 
We include a fixed effect for the type of news outlet, using General News as the reference category. To account for reporting differences across fields and variations specific to highly-publicized papers, we also include a fixed effect for the scientific subject and a random effect for each paper with 30+ pairs (all other papers are pooled in a single random effect).

\paragraph{Results.}
Compared with General News, Science \& Technology news outlets and Press Releases report findings that more closely match those from the original paper (\autoref{fig:news_outlet_type_matching_score} shows the regression coefficients). This difference likely is due to some form of audience design where the journalist is writing for a more science-savvy readership in the latter two, whereas General News journalists must more heavily paraphrase the results for lay people.



\subsection{RQ2: Comparing Social Media Accounts}

Social media play an important role in disseminating scientific findings \citep{zakhlebin2020diffusion}, so what factors affect the presentation of scientific information on social media becomes an important question. Here, we focus on the types of Twitter users who tweet about scientific findings. Based on 182K matched tweets and paper findings, we again build a linear mixed effect regression model to predict IMS. We include fixed effects of (1) if the account is run by an organization, as inferred using M3 \citep{wang2019demographic}, (2) if the account is  verified  (3) the number of followers and following, both log-transformed, and (4) the account age in years. We use the same field fixed effects and paper random effects as in RQ1.

\begin{figure}[t]
  \centering
    \includegraphics[width=0.39\textwidth]{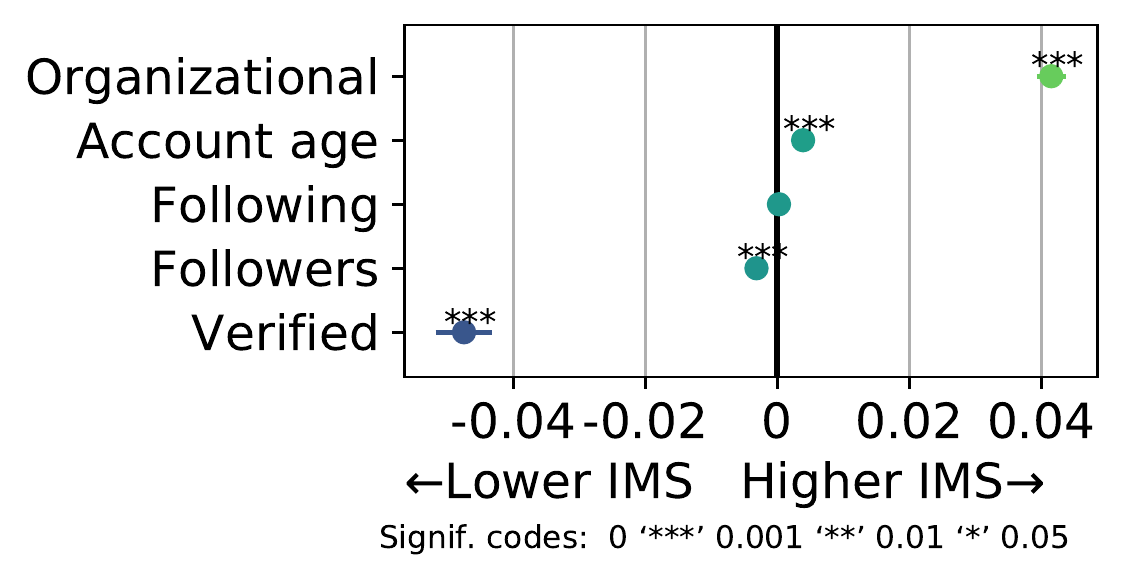}
    \caption{Organizational Twitter accounts keep more original information from the paper finding while verified users and those with more followers change more information when tweeting about a scientific finding. 
    }
    \label{fig:twitter_user_types_matching_score}
\end{figure}

\paragraph{Results}
The type of user strongly influences how faithful the tweets are to the original findings (Figure \ref{fig:twitter_user_types_matching_score}). Accounts from organizations tend to be more faithful to the original paper findings, which could be due to intentional actions of image management to build trust \citep{saffer2013effects}. Surprisingly, verified accounts were far more likely to change information away from its original meaning; similarly, accounts with more followers had the same trend. Given their prominent roles in Twitter communication \citep{bakshy2011everyone,hentschel2014finding}, multiple mechanisms may explain this gap such as adding more commentary or trying to translate original scientific findings to lay language to make the findings easier to understand. \autoref{sec:regression_details} shows the details of regression results. \cameraready{Our results point to a need to study systematic behavior based on who is communicating science.}

\subsection{RQ3: What Information Changes}
\label{sec:section_analysis}
Most studies on scientific misinformation focus on paper titles and abstracts \citep[e.g.,][]{sumner2014association}, which cannot fully reflect the information presented in the full papers. Analyzing the information change of findings paired from all sections of papers could help to better understand the mechanisms behind scientific misinformation and develop strategies to reduce them. We use the same 1.1M finding pair dataset as RQ1 and analyze what information might have changed using two models trained for changes in scientific communication: identifying exaggerations \citep{wright2021semi} and certainty \citep{DBLP:conf/emnlp/PeiJ21}. See \autoref{sec:exaggeration-supplement} for more details on the exaggeration detection task.

\begin{figure}[t]
  
  \centering
    \includegraphics[width=\linewidth]{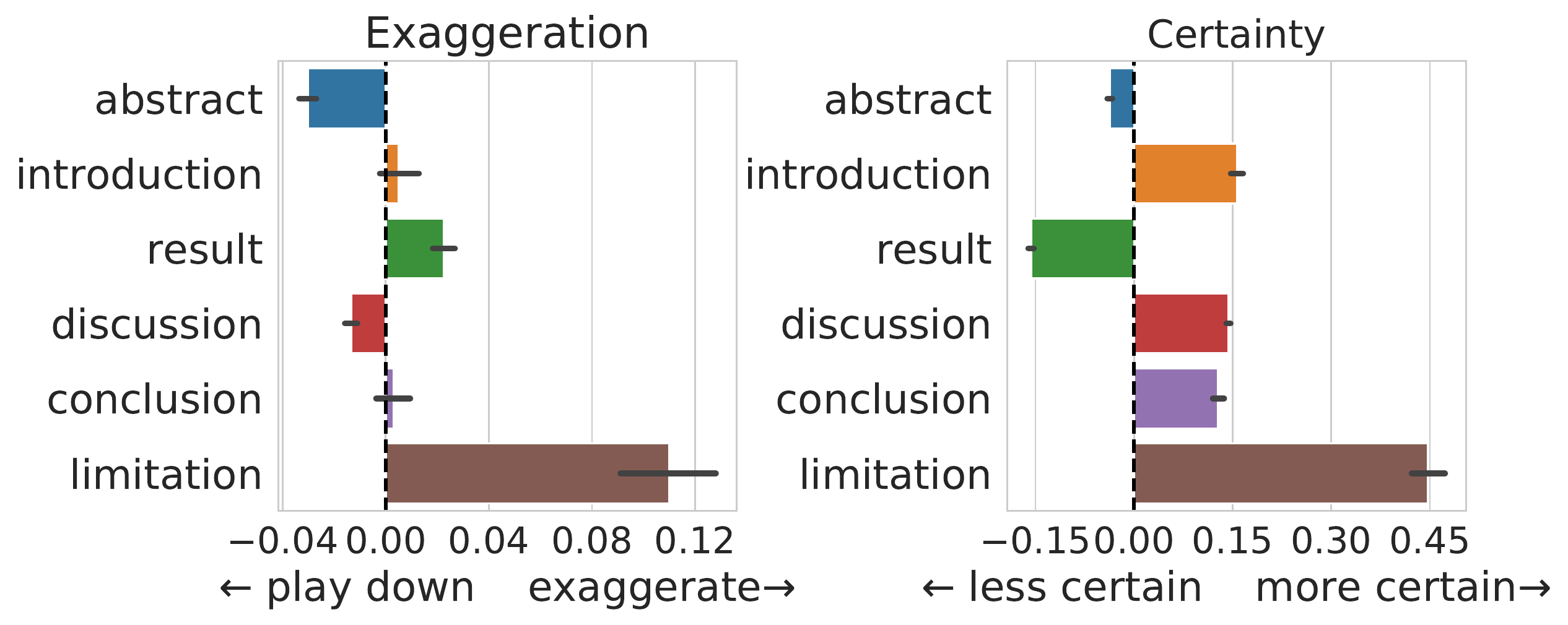}
    \caption{Journalists tend to downplay the certainty and strength of findings in abstracts, but overstate findings discussed in limitations sections. 
    }
    \label{fig:section_effect_news_paper_pairs}
\end{figure}

\paragraph{Results}
Journalists tend to downplay the certainty and strength of findings from abstracts (\autoref{fig:section_effect_news_paper_pairs}), mirroring the results of \citet{DBLP:conf/emnlp/PeiJ21}. However, this pattern does not persist for findings in other parts of papers,  especially the limitations. Existing studies suggest that journalists might fail to report the limitations of scientific findings \citep{fischhoff2012communicating}, and our results here suggest that findings presented in limitations are more likely to be exaggerated and overstated. However, it is also possible that scientists may adopt different discourse strategies for different parts of a paper \citep{clark2013writing}. Nonetheless, our result obviates the necessity of analyzing the full text of a paper when studying science communication.

\section{Conclusion}

Faithful communication of scientific results is critical for disseminating new information and establishing public trust in science. Given the challenge of---and occasional failures in---communicating science, new resources and models are needed to evaluate how science is reported. Here, we introduce \DATASET{}, a new science communication paraphrases dataset labeled with information similarity. Extensive experiments demonstrate that models can predict the degree to which two reports of a scientific finding have the same information but that this is a challenging task even for current SOTA pre-trained language models. In downstream applications, we show  \DATASET{} improves model performance for evidence retrieval for scientific fact checking; and, using the trained model to perform a large-scale analysis of information change in science communication, we show systematic behaviors in how different people and news outlets faithfully convey scientific results. Data, code, and pretrained models are available at {\small \url{http://www.copenlu.com/publication/2022_emnlp_wright/}}. 

\section*{Acknowledgements}
$\begin{array}{l}\includegraphics[width=1cm]{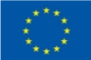} \end{array}$ This project has received funding from the European Union's Horizon 2020 research and innovation programme under the Marie Sk\l{}odowska-Curie grant agreement No 801199 and a Rackham Graduate Student Research Grant at the University of Michigan.

\section*{Limitations}
We note three limitations of our study.
%
%
Our data and analysis in social media is limited to only one platform, Twitter, and includes only tweets directly linked to the original paper, as indicated through Altmetric. While Twitter is among the largest social media platforms and is the most common in the Altmetric data, our data potentially omits other kinds of scientific communication about papers that do not directly link to a paper or tweets that link to a paper that cannot be easily identified to a DOI (e.g., linking to a PDF hosted on a personal website). Other types of tweets may be omitted from our dataset such as those written in a thread, or in a tweetorial, about a paper \cite{gero2021makes}, which may include additional tweets that describe a paper's findings. While our models would likely still be able to effectively analyze such tweets, these additional forms of scientific communication could add new variety. We leave identifying and collecting such tweets to future work. 

Second, our study focuses on only four large scientific fields. While these fields do cover a broad selection of papers, we were unable to annotate additional fields due to annotation budget and limitations from the Prolific platform. On Prolific, not all potential domains had sufficient numbers of qualified annotators (we required at least a Bachelor's degree in the domain) and the number of unique surveys to run scaled linearly with the number of domains, creating a significant human overhead. However, we will open source our annotation interface and pipeline and we encourage further efforts to build a larger dataset across more scientific domains.

Finally, while our models achieve moderately high performance at inferring the information matching (\autoref{fig:baseline-performance}), performance is not perfect, which potentially limits our ability in downstream models and tasks. While we show the data is still useful in training for related tasks (\Sref{sec:evidence-retrieval}) and a trained model can be used to identify systematic behavior by types of users and outlets (\Sref{sec:social-media-analysis}), more accurate models would likely be needed to identify any trends for finer-grained settings, such as looking at the behavior of a specific outlet. For this reason,w e have kept our analyses at a higher level (e.g., outlet categories).



\section*{Ethics and Impacts}

Miscommunication of scientific information can have negative impacts on many aspects of our society. Our study contributes to a large research program on the science of science communications \citep{national2017communicating}. Our dataset and model could be used to keep track of information change in science communication, enable large-scale analysis to understand the current science communication ecosystem, and finally help to facilitate better and more effective science communications. 


\textbf{Crowdsourcing ethics} Annotating paired findings requires deep attention and may lead to annotator burnout. We carefully designed our annotation pipeline to provide a good annotation experience for the annotators. We designed a user-friendly Web-based annotation interface that allows annotators to do annotations using keyboard shortcuts. All the annotators are encouraged to leave comments and answer several questions about their annotation experience. More than 95\% of the annotators are satisfied with their annotation experience and many people suggest that our study helps them to better understand the science communication process\footnote{For example, one participant said ``Nice learning experience, Helps to understand the news can be far more different then the research paper cited''} and our annotation interface makes their task easier.\footnote{For example, one participant said ``i liked the option of using my keyboard, it made the experience more comfortable and efficient.''}

\bibliography{anthology,custom, certainty}

\begin{thebibliography}{61}
\expandafter\ifx\csname natexlab\endcsname\relax\def\natexlab#1{#1}\fi

\bibitem[{Augenstein et~al.(2019)Augenstein, Lioma, Wang, Chaves~Lima, Hansen,
  Hansen, and Simonsen}]{DBLP:conf/emnlp/AugensteinLWLHH19}
Isabelle Augenstein, Christina Lioma, Dongsheng Wang, Lucas Chaves~Lima, Casper
  Hansen, Christian Hansen, and Jakob~Grue Simonsen. 2019.
\newblock \href {https://doi.org/10.18653/v1/D19-1475} {{M}ulti{FC}: A
  real-world multi-domain dataset for evidence-based fact checking of claims}.
\newblock In \emph{Proceedings of the 2019 Conference on Empirical Methods in
  Natural Language Processing and the 9th International Joint Conference on
  Natural Language Processing (EMNLP-IJCNLP)}, pages 4685--4697, Hong Kong,
  China. Association for Computational Linguistics.

\bibitem[{August et~al.(2020{\natexlab{a}})August, Card, Hsieh, Smith, and
  Reinecke}]{DBLP:conf/chi/AugustCHSR20}
Tal August, Dallas Card, Gary Hsieh, Noah~A. Smith, and Katharina Reinecke.
  2020{\natexlab{a}}.
\newblock \href {https://doi.org/10.1145/3313831.3376524} {Explain like {I} am
  a scientist: The linguistic barriers of entry to r/science}.
\newblock In \emph{{CHI} '20: {CHI} Conference on Human Factors in Computing
  Systems, Honolulu, HI, USA, April 25-30, 2020}, pages 1--12. {ACM}.

\bibitem[{August et~al.(2020{\natexlab{b}})August, Kim, Reinecke, and
  Smith}]{august2020writing}
Tal August, Lauren Kim, Katharina Reinecke, and Noah~A. Smith.
  2020{\natexlab{b}}.
\newblock \href {https://doi.org/10.18653/v1/2020.emnlp-main.429} {Writing
  strategies for science communication: Data and computational analysis}.
\newblock In \emph{Proceedings of the 2020 Conference on Empirical Methods in
  Natural Language Processing (EMNLP)}, pages 5327--5344, Online. Association
  for Computational Linguistics.

\bibitem[{Bakshy et~al.(2011)Bakshy, Hofman, Mason, and
  Watts}]{bakshy2011everyone}
Eytan Bakshy, Jake~M. Hofman, Winter~A. Mason, and Duncan~J. Watts. 2011.
\newblock \href {https://doi.org/10.1145/1935826.1935845} {Everyone's an
  influencer: quantifying influence on twitter}.
\newblock In \emph{Proceedings of the Forth International Conference on Web
  Search and Web Data Mining, {WSDM} 2011, Hong Kong, China, February 9-12,
  2011}, pages 65--74. {ACM}.

\bibitem[{Beltagy et~al.(2019)Beltagy, Lo, and
  Cohan}]{beltagy-etal-2019-scibert}
Iz~Beltagy, Kyle Lo, and Arman Cohan. 2019.
\newblock \href {https://doi.org/10.18653/v1/D19-1371} {{S}ci{BERT}: A
  pretrained language model for scientific text}.
\newblock In \emph{Proceedings of the 2019 Conference on Empirical Methods in
  Natural Language Processing and the 9th International Joint Conference on
  Natural Language Processing (EMNLP-IJCNLP)}, pages 3615--3620, Hong Kong,
  China. Association for Computational Linguistics.

\bibitem[{Boissonnet et~al.(2022)Boissonnet, Saeidi, Plachouras, and
  Vlachos}]{DBLP:conf/bionlp/BoissonnetSPV22}
Alodie Boissonnet, Marzieh Saeidi, Vassilis Plachouras, and Andreas Vlachos.
  2022.
\newblock \href {https://aclanthology.org/2022.bionlp-1.1} {Explainable
  assessment of healthcare articles with {QA}}.
\newblock In \emph{Proceedings of the 21st Workshop on Biomedical Language
  Processing, BioNLP@ACL 2022, Dublin, Ireland, May 26, 2022}, pages 1--9.
  Association for Computational Linguistics.

\bibitem[{Bowman et~al.(2015)Bowman, Angeli, Potts, and
  Manning}]{snli:emnlp2015}
Samuel~R. Bowman, Gabor Angeli, Christopher Potts, and Christopher~D. Manning.
  2015.
\newblock \href {https://doi.org/10.18653/v1/D15-1075} {A large annotated
  corpus for learning natural language inference}.
\newblock In \emph{Proceedings of the 2015 Conference on Empirical Methods in
  Natural Language Processing}, pages 632--642, Lisbon, Portugal. Association
  for Computational Linguistics.

\bibitem[{Bratton et~al.(2019)Bratton, Adams, Challenger, Boivin, Bott,
  Chambers, and Sumner}]{bratton2019association}
Luke Bratton, Rachel~C Adams, Aim{\'e}e Challenger, Jacky Boivin, Lewis Bott,
  Christopher~D Chambers, and Petroc Sumner. 2019.
\newblock {The Association Between Exaggeration in Health-Related Science News
  and Academic Press Releases: A Replication Study}.
\newblock \emph{Wellcome open research}, 4.

\bibitem[{Canese and Weis(2013)}]{canese2013pubmed}
Kathi Canese and Sarah Weis. 2013.
\newblock Pubmed: the bibliographic database.
\newblock \emph{The NCBI handbook}, 2(1).

\bibitem[{Cer et~al.(2017)Cer, Diab, Agirre, Lopez-Gazpio, and
  Specia}]{DBLP:journals/corr/abs-1708-00055}
Daniel Cer, Mona Diab, Eneko Agirre, I{\~n}igo Lopez-Gazpio, and Lucia Specia.
  2017.
\newblock \href {https://doi.org/10.18653/v1/S17-2001} {{S}em{E}val-2017 task
  1: Semantic textual similarity multilingual and crosslingual focused
  evaluation}.
\newblock In \emph{Proceedings of the 11th International Workshop on Semantic
  Evaluation ({S}em{E}val-2017)}, pages 1--14, Vancouver, Canada. Association
  for Computational Linguistics.

\bibitem[{Clark(2013)}]{clark2013writing}
Sarah~Kartchner Clark. 2013.
\newblock \emph{Writing strategies for science}.
\newblock Teacher Created Materials.

\bibitem[{Cohan et~al.(2020)Cohan, Feldman, Beltagy, Downey, and
  Weld}]{DBLP:conf/acl/CohanFBDW20}
Arman Cohan, Sergey Feldman, Iz~Beltagy, Doug Downey, and Daniel Weld. 2020.
\newblock \href {https://doi.org/10.18653/v1/2020.acl-main.207} {{SPECTER}:
  Document-level representation learning using citation-informed transformers}.
\newblock In \emph{Proceedings of the 58th Annual Meeting of the Association
  for Computational Linguistics}, pages 2270--2282, Online. Association for
  Computational Linguistics.

\bibitem[{Dai et~al.(2020)Dai, Sun, and Wang}]{DBLP:conf/icwsm/DaiSW20}
Enyan Dai, Yiwei Sun, and Suhang Wang. 2020.
\newblock \href {https://ojs.aaai.org/index.php/ICWSM/article/view/7350}
  {Ginger cannot cure cancer: Battling fake health news with a comprehensive
  data repository}.
\newblock In \emph{Proceedings of the Fourteenth International {AAAI}
  Conference on Web and Social Media, {ICWSM} 2020, Held Virtually, Original
  Venue: Atlanta, Georgia, USA, June 8-11, 2020}, pages 853--862. {AAAI} Press.

\bibitem[{Fang et~al.(2016)Fang, Wang, Han, He, Wei, Zhao, Imam, Ping, Li, Xu
  et~al.}]{fang2016dietary}
Xuexian Fang, Kai Wang, Dan Han, Xuyan He, Jiayu Wei, Lu~Zhao, Mustapha~Umar
  Imam, Zhiguang Ping, Yusheng Li, Yuming Xu, et~al. 2016.
\newblock Dietary magnesium intake and the risk of cardiovascular disease, type
  2 diabetes, and all-cause mortality: a dose--response meta-analysis of
  prospective cohort studies.
\newblock \emph{BMC medicine}, 14(1):1--13.

\bibitem[{Fischhoff(2012)}]{fischhoff2012communicating}
Baruch Fischhoff. 2012.
\newblock Communicating uncertainty fulfilling the duty to inform.
\newblock \emph{Issues in Science and Technology}, 28(4):63--70.

\bibitem[{Ga{\l}ecki and Burzykowski(2013)}]{galecki2013linear}
Andrzej Ga{\l}ecki and Tomasz Burzykowski. 2013.
\newblock Linear mixed-effects model.
\newblock In \emph{Linear mixed-effects models using R}, pages 245--273.
  Springer.

\bibitem[{Ganitkevitch et~al.(2013)Ganitkevitch, Van~Durme, and
  Callison-Burch}]{DBLP:conf/naacl/GanitkevitchDC13}
Juri Ganitkevitch, Benjamin Van~Durme, and Chris Callison-Burch. 2013.
\newblock \href {https://aclanthology.org/N13-1092} {{PPDB}: The paraphrase
  database}.
\newblock In \emph{Proceedings of the 2013 Conference of the North {A}merican
  Chapter of the Association for Computational Linguistics: Human Language
  Technologies}, pages 758--764, Atlanta, Georgia. Association for
  Computational Linguistics.

\bibitem[{Gero et~al.(2021)Gero, Liu, Huang, Lee, and Chilton}]{gero2021makes}
Katy~Ilonka Gero, Vivian Liu, Sarah Huang, Jennifer Lee, and Lydia~B Chilton.
  2021.
\newblock What makes tweetorials tick: How experts communicate complex topics
  on twitter.
\newblock \emph{Proceedings of the ACM on Human-Computer Interaction},
  5(CSCW2):1--26.

\bibitem[{Ginev and Miller(2020)}]{DBLP:conf/lrec/GinevM20}
Deyan Ginev and Bruce~R Miller. 2020.
\newblock \href {https://aclanthology.org/2020.lrec-1.153} {Scientific
  statement classification over ar{X}iv.org}.
\newblock In \emph{Proceedings of the 12th Language Resources and Evaluation
  Conference}, pages 1219--1226, Marseille, France. European Language Resources
  Association.

\bibitem[{Gustafson and Rice(2019)}]{gustafson2019effects}
Abel Gustafson and Ronald~E Rice. 2019.
\newblock The effects of uncertainty frames in three science communication
  topics.
\newblock \emph{Science Communication}, 41(6):679--706.

\bibitem[{Hentschel et~al.(2014)Hentschel, Alonso, Counts, and
  Kandylas}]{hentschel2014finding}
Martin Hentschel, Omar Alonso, Scott Counts, and Vasileios Kandylas. 2014.
\newblock Finding users we trust: Scaling up verified twitter users using their
  communication patterns.
\newblock In \emph{Eighth International AAAI Conference on Weblogs and Social
  Media}.

\bibitem[{Holub et~al.(2008)Holub, Perona, and Burl}]{DBLP:conf/cvpr/HolubPB08}
Alex Holub, Pietro Perona, and Michael~C. Burl. 2008.
\newblock \href {https://doi.org/10.1109/CVPRW.2008.4563068} {Entropy-based
  active learning for object recognition}.
\newblock In \emph{{IEEE} Conference on Computer Vision and Pattern
  Recognition, {CVPR} Workshops 2008, Anchorage, AK, USA, 23-28 June, 2008},
  pages 1--8. {IEEE} Computer Society.

\bibitem[{Houlsby et~al.(2011)Houlsby, Huszar, Ghahramani, and
  Lengyel}]{DBLP:journals/corr/abs-1112-5745}
Neil Houlsby, Ferenc Huszar, Zoubin Ghahramani, and M{\'{a}}t{\'{e}} Lengyel.
  2011.
\newblock \href {http://arxiv.org/abs/1112.5745} {Bayesian active learning for
  classification and preference learning}.
\newblock \emph{CoRR}, abs/1112.5745.

\bibitem[{Hovy et~al.(2013)Hovy, Berg-Kirkpatrick, Vaswani, and
  Hovy}]{DBLP:conf/naacl/HovyBVH13}
Dirk Hovy, Taylor Berg-Kirkpatrick, Ashish Vaswani, and Eduard Hovy. 2013.
\newblock \href {https://aclanthology.org/N13-1132} {Learning whom to trust
  with {MACE}}.
\newblock In \emph{Proceedings of the 2013 Conference of the North {A}merican
  Chapter of the Association for Computational Linguistics: Human Language
  Technologies}, pages 1120--1130, Atlanta, Georgia. Association for
  Computational Linguistics.

\bibitem[{Kuru et~al.(2021)Kuru, Stecula, Lu, Ophir, Chan, Winneg,
  Hall~Jamieson, and Albarrac{\'\i}n}]{kuru2021effects}
Ozan Kuru, Dominik Stecula, Hang Lu, Yotam Ophir, Man-pui~Sally Chan, Ken
  Winneg, Kathleen Hall~Jamieson, and Dolores Albarrac{\'\i}n. 2021.
\newblock The effects of scientific messages and narratives about vaccination.
\newblock \emph{PLoS One}, 16(3):e0248328.

\bibitem[{Liu et~al.(2019)Liu, Ott, Goyal, Du, Joshi, Chen, Levy, Lewis,
  Zettlemoyer, and Stoyanov}]{DBLP:journals/corr/abs-1907-11692}
Yinhan Liu, Myle Ott, Naman Goyal, Jingfei Du, Mandar Joshi, Danqi Chen, Omer
  Levy, Mike Lewis, Luke Zettlemoyer, and Veselin Stoyanov. 2019.
\newblock \href {http://arxiv.org/abs/1907.11692} {Roberta: {A} robustly
  optimized {BERT} pretraining approach}.
\newblock \emph{CoRR}, abs/1907.11692.

\bibitem[{Lo et~al.(2020)Lo, Wang, Neumann, Kinney, and
  Weld}]{DBLP:conf/acl/LoWNKW20}
Kyle Lo, Lucy~Lu Wang, Mark Neumann, Rodney Kinney, and Daniel Weld. 2020.
\newblock \href {https://doi.org/10.18653/v1/2020.acl-main.447} {{S}2{ORC}: The
  semantic scholar open research corpus}.
\newblock In \emph{Proceedings of the 58th Annual Meeting of the Association
  for Computational Linguistics}, pages 4969--4983, Online. Association for
  Computational Linguistics.

\bibitem[{MacKay(1992)}]{DBLP:journals/neco/MacKay92b}
David J.~C. MacKay. 1992.
\newblock \href {https://doi.org/10.1162/neco.1992.4.4.590} {Information-based
  objective functions for active data selection}.
\newblock \emph{Neural Comput.}, 4(4):590--604.

\bibitem[{McCarthy and Jarvis(2010)}]{mccarthy2010mtld}
Philip~M McCarthy and Scott Jarvis. 2010.
\newblock Mtld, vocd-d, and hd-d: A validation study of sophisticated
  approaches to lexical diversity assessment.
\newblock \emph{Behavior research methods}, 42(2):381--392.

\bibitem[{Mencher and Shilton(1997)}]{mencher1997news}
Melvin Mencher and Wendy~P Shilton. 1997.
\newblock \emph{News reporting and writing}.
\newblock Brown \& Benchmark Publishers Madison, WI.

\bibitem[{Mohr et~al.(2022)Mohr, W{\"{u}}hrl, and
  Klinger}]{DBLP:journals/corr/abs-2204-12164}
Isabelle Mohr, Amelie W{\"{u}}hrl, and Roman Klinger. 2022.
\newblock \href {https://doi.org/10.48550/arXiv.2204.12164} {Covert: {A} corpus
  of fact-checked biomedical {COVID-19} tweets}.
\newblock \emph{CoRR}, abs/2204.12164.

\bibitem[{{National Academies of Sciences, Engineering, and
  Medicine}(2017)}]{national2017communicating}
{National Academies of Sciences, Engineering, and Medicine}. 2017.
\newblock Communicating science effectively: A research agenda.

\bibitem[{Nie et~al.(2020)Nie, Williams, Dinan, Bansal, Weston, and
  Kiela}]{nie-etal-2020-adversarial}
Yixin Nie, Adina Williams, Emily Dinan, Mohit Bansal, Jason Weston, and Douwe
  Kiela. 2020.
\newblock \href {https://doi.org/10.18653/v1/2020.acl-main.441} {Adversarial
  {NLI}: A new benchmark for natural language understanding}.
\newblock In \emph{Proceedings of the 58th Annual Meeting of the Association
  for Computational Linguistics}, pages 4885--4901, Online. Association for
  Computational Linguistics.

\bibitem[{Nighojkar and Licato(2021)}]{nighojkar-licato-2021-improving}
Animesh Nighojkar and John Licato. 2021.
\newblock \href {https://doi.org/10.18653/v1/2021.acl-long.552} {Improving
  paraphrase detection with the adversarial paraphrasing task}.
\newblock In \emph{Proceedings of the 59th Annual Meeting of the Association
  for Computational Linguistics and the 11th International Joint Conference on
  Natural Language Processing (Volume 1: Long Papers)}, pages 7106--7116,
  Online. Association for Computational Linguistics.

\bibitem[{Pei et~al.(2022)Pei, Ananthasubramaniam, Wang, Zhou, Dedeloudis,
  Sargent, and Jurgens}]{pei2022potato}
Jiaxin Pei, Aparna Ananthasubramaniam, Xingyao Wang, Naitian Zhou, Apostolos
  Dedeloudis, Jackson Sargent, and David Jurgens. 2022.
\newblock Potato: The portable text annotation tool.
\newblock In \emph{Proceedings of the 2022 Conference on Empirical Methods in
  Natural Language Processing: System Demonstrations}.

\bibitem[{Pei and Jurgens(2021)}]{DBLP:conf/emnlp/PeiJ21}
Jiaxin Pei and David Jurgens. 2021.
\newblock \href {https://doi.org/10.18653/v1/2021.emnlp-main.784} {Measuring
  sentence-level and aspect-level (un)certainty in science communications}.
\newblock In \emph{Proceedings of the 2021 Conference on Empirical Methods in
  Natural Language Processing, {EMNLP} 2021, Virtual Event / Punta Cana,
  Dominican Republic, 7-11 November, 2021}, pages 9959--10011. Association for
  Computational Linguistics.

\bibitem[{Prabhakaran et~al.(2016)Prabhakaran, Hamilton, McFarland, and
  Jurafsky}]{Prabhakaran2016PredictingTR}
Vinodkumar Prabhakaran, William~L. Hamilton, Dan McFarland, and Dan Jurafsky.
  2016.
\newblock \href {https://doi.org/10.18653/v1/P16-1111} {Predicting the rise and
  fall of scientific topics from trends in their rhetorical framing}.
\newblock In \emph{Proceedings of the 54th Annual Meeting of the Association
  for Computational Linguistics (Volume 1: Long Papers)}, pages 1170--1180,
  Berlin, Germany. Association for Computational Linguistics.

\bibitem[{Reimers and Gurevych(2019)}]{reimers-2019-sentence-bert}
Nils Reimers and Iryna Gurevych. 2019.
\newblock \href {https://doi.org/10.18653/v1/D19-1410} {Sentence-{BERT}:
  Sentence embeddings using {S}iamese {BERT}-networks}.
\newblock In \emph{Proceedings of the 2019 Conference on Empirical Methods in
  Natural Language Processing and the 9th International Joint Conference on
  Natural Language Processing (EMNLP-IJCNLP)}, pages 3982--3992, Hong Kong,
  China. Association for Computational Linguistics.

\bibitem[{Richardson(1990)}]{richardson1990writing}
Laurel Richardson. 1990.
\newblock \emph{Writing strategies: Reaching diverse audiences}, volume~21.
\newblock Sage Publications.

\bibitem[{Robertson et~al.(1994)Robertson, Walker, Jones, Hancock-Beaulieu, and
  Gatford}]{Robertson1994OkapiAT}
Stephen~E. Robertson, Steve Walker, Susan Jones, Micheline Hancock-Beaulieu,
  and Mike Gatford. 1994.
\newblock Okapi at trec-3.
\newblock In \emph{TREC}.

\bibitem[{Roland(2009)}]{roland2009quality}
Marie-Claude Roland. 2009.
\newblock Quality and integrity in scientific writing: prerequisites for
  quality in science communication.
\newblock \emph{Journal of Science Communication}, 8(2):A04.

\bibitem[{Saakyan et~al.(2021)Saakyan, Chakrabarty, and
  Muresan}]{DBLP:conf/acl/SaakyanCM20}
Arkadiy Saakyan, Tuhin Chakrabarty, and Smaranda Muresan. 2021.
\newblock \href {https://doi.org/10.18653/v1/2021.acl-long.165} {{COVID}-fact:
  Fact extraction and verification of real-world claims on {COVID}-19
  pandemic}.
\newblock In \emph{Proceedings of the 59th Annual Meeting of the Association
  for Computational Linguistics and the 11th International Joint Conference on
  Natural Language Processing (Volume 1: Long Papers)}, pages 2116--2129,
  Online. Association for Computational Linguistics.

\bibitem[{Saffer et~al.(2013)Saffer, Sommerfeldt, and
  Taylor}]{saffer2013effects}
Adam~J Saffer, Erich~J Sommerfeldt, and Maureen Taylor. 2013.
\newblock The effects of organizational twitter interactivity on
  organization--public relationships.
\newblock \emph{Public relations review}, 39(3):213--215.

\bibitem[{Salita(2015)}]{Salita2015WritingFL}
Joselita~T. Salita. 2015.
\newblock Writing for lay audiences: A challenge for scientists.
\newblock \emph{{Medical Writing}}, 24:183--189.

\bibitem[{Shaar et~al.(2020)Shaar, Babulkov, Da~San~Martino, and
  Nakov}]{shaar-etal-2020-known}
Shaden Shaar, Nikolay Babulkov, Giovanni Da~San~Martino, and Preslav Nakov.
  2020.
\newblock \href {https://doi.org/10.18653/v1/2020.acl-main.332} {That is a
  known lie: Detecting previously fact-checked claims}.
\newblock In \emph{Proceedings of the 58th Annual Meeting of the Association
  for Computational Linguistics}, pages 3607--3618, Online. Association for
  Computational Linguistics.

\bibitem[{Song et~al.(2020)Song, Tan, Qin, Lu, and
  Liu}]{DBLP:conf/nips/Song0QLL20}
Kaitao Song, Xu~Tan, Tao Qin, Jianfeng Lu, and Tie{-}Yan Liu. 2020.
\newblock \href
  {https://proceedings.neurips.cc/paper/2020/hash/c3a690be93aa602ee2dc0ccab5b7b67e-Abstract.html}
  {Mpnet: Masked and permuted pre-training for language understanding}.
\newblock In \emph{Advances in Neural Information Processing Systems 33: Annual
  Conference on Neural Information Processing Systems 2020, NeurIPS 2020,
  December 6-12, 2020, virtual}.

\bibitem[{Stolaroff et~al.(2018)Stolaroff, Samaras, O’Neill, Lubers,
  Mitchell, and Ceperley}]{stolaroff2018energy}
Joshuah~K Stolaroff, Constantine Samaras, Emma~R O’Neill, Alia Lubers,
  Alexandra~S Mitchell, and Daniel Ceperley. 2018.
\newblock Energy use and life cycle greenhouse gas emissions of drones for
  commercial package delivery.
\newblock \emph{Nature communications}, 9(1):1--13.

\bibitem[{Sumner et~al.(2014)Sumner, Vivian-Griffiths, Boivin, Williams,
  Venetis, Davies, Ogden, Whelan, Hughes, Dalton
  et~al.}]{sumner2014association}
Petroc Sumner, Solveiga Vivian-Griffiths, Jacky Boivin, Andy Williams,
  Christos~A Venetis, Aim{\'e}e Davies, Jack Ogden, Leanne Whelan, Bethan
  Hughes, Bethan Dalton, et~al. 2014.
\newblock {The Association Between Exaggeration in Health Related Science News
  and Academic Press Releases: Retrospective Observational Study}.
\newblock \emph{BMJ}, 349.

\bibitem[{Tan and Lee(2014)}]{DBLP:conf/acl/TanL14}
Chenhao Tan and Lillian Lee. 2014.
\newblock \href {https://doi.org/10.3115/v1/P14-2066} {A corpus of
  sentence-level revisions in academic writing: A step towards understanding
  statement strength in communication}.
\newblock In \emph{Proceedings of the 52nd Annual Meeting of the Association
  for Computational Linguistics (Volume 2: Short Papers)}, pages 403--408,
  Baltimore, Maryland. Association for Computational Linguistics.

\bibitem[{Thorne et~al.(2018)Thorne, Vlachos, Christodoulopoulos, and
  Mittal}]{DBLP:conf/naacl/ThorneVCM18}
James Thorne, Andreas Vlachos, Christos Christodoulopoulos, and Arpit Mittal.
  2018.
\newblock \href {https://doi.org/10.18653/v1/N18-1074} {{FEVER}: a large-scale
  dataset for fact extraction and {VER}ification}.
\newblock In \emph{Proceedings of the 2018 Conference of the North {A}merican
  Chapter of the Association for Computational Linguistics: Human Language
  Technologies, Volume 1 (Long Papers)}, pages 809--819, New Orleans,
  Louisiana. Association for Computational Linguistics.

\bibitem[{Vadapalli et~al.(2018)Vadapalli, Syed, Prabhu, Srinivasan, and
  Varma}]{DBLP:conf/emnlp/VadapalliSPSV18}
Raghuram Vadapalli, Bakhtiyar Syed, Nishant Prabhu, Balaji~Vasan Srinivasan,
  and Vasudeva Varma. 2018.
\newblock \href {https://doi.org/10.18653/v1/D18-2028} {When science journalism
  meets artificial intelligence : An interactive demonstration}.
\newblock In \emph{Proceedings of the 2018 Conference on Empirical Methods in
  Natural Language Processing: System Demonstrations}, pages 163--168,
  Brussels, Belgium. Association for Computational Linguistics.

\bibitem[{Wadden et~al.(2020)Wadden, Lin, Lo, Wang, van Zuylen, Cohan, and
  Hajishirzi}]{DBLP:conf/emnlp/WaddenLLWZCH20}
David Wadden, Shanchuan Lin, Kyle Lo, Lucy~Lu Wang, Madeleine van Zuylen, Arman
  Cohan, and Hannaneh Hajishirzi. 2020.
\newblock \href {https://doi.org/10.18653/v1/2020.emnlp-main.609} {Fact or
  fiction: Verifying scientific claims}.
\newblock In \emph{Proceedings of the 2020 Conference on Empirical Methods in
  Natural Language Processing (EMNLP)}, pages 7534--7550, Online. Association
  for Computational Linguistics.

\bibitem[{Wang et~al.(2020{\natexlab{a}})Wang, Wei, Dong, Bao, Yang, and
  Zhou}]{DBLP:conf/nips/WangW0B0020}
Wenhui Wang, Furu Wei, Li~Dong, Hangbo Bao, Nan Yang, and Ming Zhou.
  2020{\natexlab{a}}.
\newblock \href
  {https://proceedings.neurips.cc/paper/2020/hash/3f5ee243547dee91fbd053c1c4a845aa-Abstract.html}
  {Minilm: Deep self-attention distillation for task-agnostic compression of
  pre-trained transformers}.
\newblock In \emph{Advances in Neural Information Processing Systems 33: Annual
  Conference on Neural Information Processing Systems 2020, NeurIPS 2020,
  December 6-12, 2020, virtual}.

\bibitem[{Wang et~al.(2020{\natexlab{b}})Wang, Wei, Dong, Bao, Yang, and
  Zhou}]{wang2020minilm}
Wenhui Wang, Furu Wei, Li~Dong, Hangbo Bao, Nan Yang, and Ming Zhou.
  2020{\natexlab{b}}.
\newblock \href {http://arxiv.org/abs/2002.10957} {Minilm: Deep self-attention
  distillation for task-agnostic compression of pre-trained transformers}.

\bibitem[{Wang et~al.(2019)Wang, Hale, Adelani, Grabowicz, Hartmann,
  Fl{\"{o}}ck, and Jurgens}]{wang2019demographic}
Zijian Wang, Scott~A. Hale, David~Ifeoluwa Adelani, Przemyslaw~A. Grabowicz,
  Timo Hartmann, Fabian Fl{\"{o}}ck, and David Jurgens. 2019.
\newblock \href {https://doi.org/10.1145/3308558.3313684} {Demographic
  inference and representative population estimates from multilingual social
  media data}.
\newblock In \emph{The World Wide Web Conference, {WWW} 2019, San Francisco,
  CA, USA, May 13-17, 2019}, pages 2056--2067. {ACM}.

\bibitem[{Weigold(2001)}]{weigold2001communicating}
Michael~F Weigold. 2001.
\newblock Communicating science: A review of the literature.
\newblock \emph{Science communication}, 23(2):164--193.

\bibitem[{Williams et~al.(2018)Williams, Nangia, and Bowman}]{N18-1101}
Adina Williams, Nikita Nangia, and Samuel Bowman. 2018.
\newblock \href {https://doi.org/10.18653/v1/N18-1101} {A broad-coverage
  challenge corpus for sentence understanding through inference}.
\newblock In \emph{Proceedings of the 2018 Conference of the North {A}merican
  Chapter of the Association for Computational Linguistics: Human Language
  Technologies, Volume 1 (Long Papers)}, pages 1112--1122, New Orleans,
  Louisiana. Association for Computational Linguistics.

\bibitem[{Wright and Augenstein(2021{\natexlab{a}})}]{wright2021citeworth}
Dustin Wright and Isabelle Augenstein. 2021{\natexlab{a}}.
\newblock \href {https://doi.org/10.18653/v1/2021.findings-acl.157}
  {{C}ite{W}orth: Cite-worthiness detection for improved scientific document
  understanding}.
\newblock In \emph{Findings of the Association for Computational Linguistics:
  ACL-IJCNLP 2021}, pages 1796--1807, Online. Association for Computational
  Linguistics.

\bibitem[{Wright and Augenstein(2021{\natexlab{b}})}]{wright2021semi}
Dustin Wright and Isabelle Augenstein. 2021{\natexlab{b}}.
\newblock Semi-supervised exaggeration detection of health science press
  releases.
\newblock In \emph{Proceedings of the 2021 Conference on Empirical Methods in
  Natural Language Processing}, pages 10824--10836.

\bibitem[{Wright et~al.(2022)Wright, Wadden, Lo, Kuehl, Cohan, Augenstein, and
  Wang}]{DBLP:conf/acl/0001WLKCAW22}
Dustin Wright, David Wadden, Kyle Lo, Bailey Kuehl, Arman Cohan, Isabelle
  Augenstein, and Lucy~Lu Wang. 2022.
\newblock \href {https://aclanthology.org/2022.acl-long.175} {Generating
  scientific claims for zero-shot scientific fact checking}.
\newblock In \emph{Proceedings of the 60th Annual Meeting of the Association
  for Computational Linguistics (Volume 1: Long Papers), {ACL} 2022, Dublin,
  Ireland, May 22-27, 2022}, pages 2448--2460. Association for Computational
  Linguistics.

\bibitem[{Zakhlebin and Horv{\'a}t(2020)}]{zakhlebin2020diffusion}
Igor Zakhlebin and Emoke-Agnes Horv{\'a}t. 2020.
\newblock Diffusion of scientific articles across online platforms.
\newblock In \emph{Proceedings of the International AAAI Conference on Web and
  Social Media}, volume~14, pages 762--773.

\end{thebibliography}
\bibliographystyle{acl_natbib}

\clearpage
\appendix

\section{Information Change vs. Semantic Similarity}
\label{sec:info-change-supplement}
\begin{table*}[t]
\small

\newcommand{\tabincell}[2]{\begin{tabular}{@{}#1@{}}#2\end{tabular}}
\resizebox{0.99\textwidth}{!}{
\begin{tabular}{p{60mm}p{60mm}}
\toprule
\textbf{Sentence 1} & \textbf{Sentence 2}\\
\midrule
The polar bear is sliding on the snow. & A polar bear is sliding across the snow.
\\
\midrule
A plane is taking off & An air plane is taking off\\
\midrule
A dog rides a skateboard & A dog is riding a skateboard\\
\midrule
A man is playing the drums & A man plays the drum\\
\bottomrule
\end{tabular}
}
\caption{Samples of sentence pairs in STSB which have a similarity score of 5 }
\label{tab:stsb-5-examples}
\end{table*}

\begin{table*}[t]
\small

\newcommand{\tabincell}[2]{\begin{tabular}{@{}#1@{}}#2\end{tabular}}
\resizebox{0.99\textwidth}{!}{
\begin{tabular}{p{60mm}p{60mm}}
\toprule
\textbf{Sentence 1} & \textbf{Sentence 2}\\
\midrule
Higher-income professionals had less tolerance for smartphone use in business meetings. & We are intrigued by the result that professionals with higher incomes are less accepting of mobile phone use in meetings.
\\
\midrule
If we allow people to retract recently posted comments, then we may be able to minimize regret from posting in the heat of the moment. & Allowing users to retract recently posted comments may help minimize regret .\\
\midrule
Papers with shorter titles get more citations \#science \#metascience \#sciencemetrics & Our analysis suggests that papers with shorter titles do receive greater numbers of citations.\\
\midrule
Low levels of self-esteem and poor emotional processing skills were significantly correlated with gang involvement, as were low levels of parental monitoring, poor parental communication and housing instability. & Major findings also indicated that low levels of parental monitoring, poor parental communication and housing instability were significantly associated with gang involvement.\\
\bottomrule
\end{tabular}
}
\caption{Samples of sentence pairs in \DATASET{} which have an \SCORE{} of 5.}
\label{tab:dataset-5-examples}
\end{table*}
We wish to highlight key differences between information change and semantic similarity, particularly with an eye to what makes the task introduced in \DATASET{} difficult compared to semantic similarity scoring. To illustrate this, we present a sample of pairs in STSB that have the highest similarity score of `5' vs. samples in \DATASET{} which have an \SCORE{} of 5 in \autoref{tab:stsb-5-examples} and \autoref{tab:dataset-5-examples}.

In this, for a pair to be perfectly similar from a semantics perspective, the entire sentence must contain exactly equivalent meaning. This is not the case with our task. For the information change task, pairs are highly similar even if some aspects of the semantics of the sentence are changed e.g. in the first sample, there is a difference between the two sentences semantically: the second in the pair discusses ``being intrigued'' by the finding, which is shared between the pair. This also makes the task extremely difficult -- a model must learn to compare only the salient scientific facts between the pair of sentences, as opposed to the entire meaning of each sentence.

\section{Pilot Annotation Details}
\label{sec:pilot-annotation-details}
\begin{figure}[t]
  
  \centering
    \includegraphics[width=\linewidth]{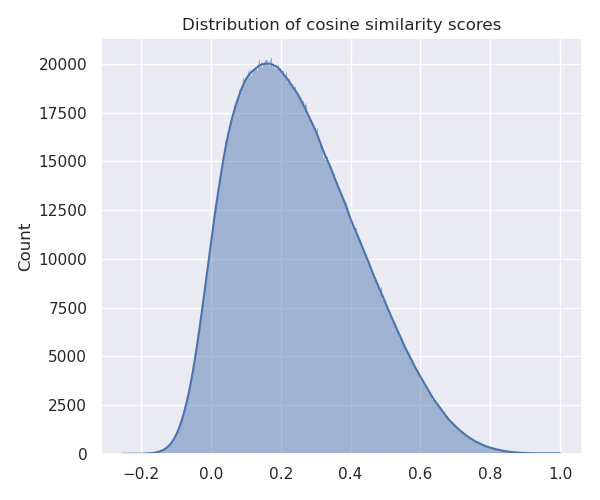}
    \caption{Distribution of the cosine similarity between findings extracted from news articles about particular scientific papers. Cosine similarity is measured between the embeddings produced for both findings using SBERT~\cite{reimers-2019-sentence-bert}.} 
    \label{fig:sample-cos-distribution}
\end{figure}

\begin{figure}[t]
  
  \centering
    \includegraphics[width=\linewidth]{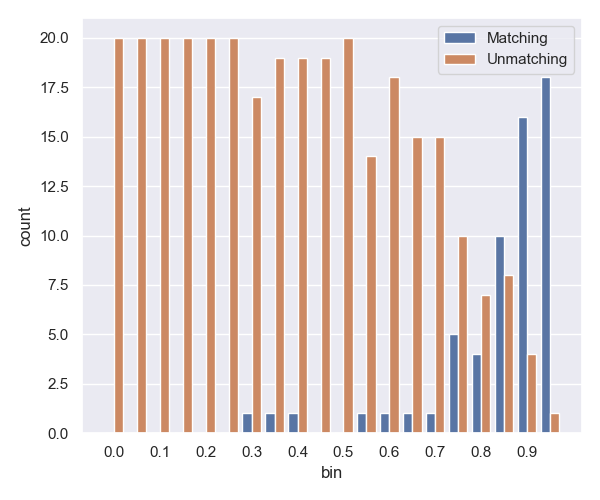}
    \caption{Number of samples per bin rated as matching vs. not matching (samples limited to those where both annotators agreed on the label). Most matching samples come from higher similarity bins, while more difficult samples come from the middle bins. 
    }
    \label{fig:pos-samples-pilot}
\end{figure}
For the pilot, we use 20 pairs from 20 different cosine similarity score bins in increments of $0.05$ starting from $0$. In other words, we have 20 bins with ranges of scores as: $0.0-0.05$, $0.05-0.1$...$0.9-0.95$, $0.95-1.0$. This results in 400 samples to annotate. 
The score distribution from 7,392,690 pairs from 3,525 source papers which we use for sampling is given in~\autoref{fig:sample-cos-distribution}. 
Each sample is annotated by two of the authors of the study with a binary label of ``matching'' vs ``not matching'', yielding a Krippendorff's alpha of $0.73$.

The number of positive samples per bin from the pilot study is given in \autoref{fig:pos-samples-pilot}. We see here that bins with a cosine similarity below 0.65 tend to have very few positive samples, and only above 0.8 do we start to see many positive samples in the bins. Almost all samples above $0.9$ are matching, and the only unmatched pairs appear to be instances of SBERT failing, since the matched pairs are almost exactly copied text. Additionally, this histogram indicates that the base rate of positive matching findings is low as the overall distribution of samples in the high cosine similarity region, where most of the matches exist, is small. At the same time, we note that some of the matches we find in the lower cosine similarity regions constitute quite interesting samples; for example, the following which has a cosine similarity of 0.41.

\begin{quote}
    \textbf{Paper finding:} For cases comparing a drone and a vehicle carrying a single package over similar distances, for example, a customer picking up a package from a retail store, the drone is clearly a lower-impact solution.~\cite{stolaroff2018energy}
    
    \textbf{News finding:} But if you forgot that essential ingredient for tonight's dinner, our findings suggest it's much better to have the grocery store send it to you by drone rather than to take your car to the store and back.\footnote{https://www.enbridge.com/energy-matters/news-and-views/delivering-packages-with-drones-might-be-good-for-the-environment}
\end{quote}

Both sentences are talking about the same finding, that drone delivery is more efficient over short distances than using a car, but in entirely different ways. From this, it is clear that simply using semantic text similarity is insufficient for solving this task, and we should include some of these lower similarity samples in our annotation. We, therefore, propose the following sampling scheme in order to balance the number of annotations we can acquire, the yield of positive samples, and the sample difficulty:
\begin{itemize}[noitemsep,nolistsep]
    \item Label all samples with a cosine similarity below $0.4$ as unmatched.
    \item Label all samples above $0.9$ with a Jaccard index above $0.5$ as matching.
    \item Sample an equal number of pairs from each $0.05$ increment bin between $0.4$ and $0.9$ for human expert annotation.
\end{itemize}

\section{Experimented annotation}
\label{sec:experimented-annotations-supplement}
We experimented with two annotation schemas: a binary schema where the annotators are asked to label ``whether the two sentences are discussing the same scientific finding'' with Yes or No, and a Likert schema where the annotators are asked to label if
%
``The information in the findings is...''
\begin{itemize}[noitemsep,nolistsep]
    \item 1: Completely different
    \item 2: Mostly different
    \item 3: Somewhat similar
    \item 4: Mostly the same
    \item 5: Completely the same
\end{itemize}
We ran several pilots using the two annotation schemas and the Likert a schema led to higher inter-annotator agreement (0.45 Krippendorff's alpha) compared with the binary schema (0.21 Krippendorff's alpha). Therefore we adopt the 5-point Likert schema for the annotation.

\section{Full Annotation Instructions}
\label{sec:annotation-instructions}
\begin{figure*}[t]
  
  \centering
    \includegraphics[width=\linewidth]{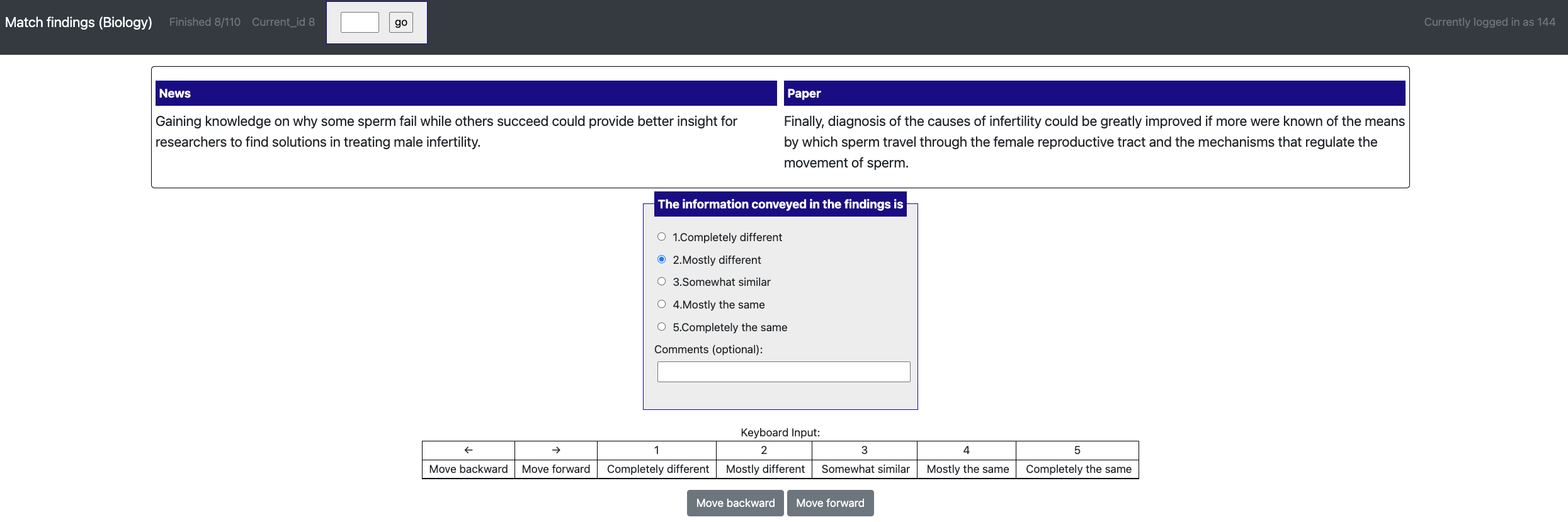}
    \caption{The annotation page of our crowdsourcing task
    }
    \label{fig:annotation_interface}
\end{figure*}

Annotation was performed using Prolific workers who labeled using \textsc{Potato} \cite{pei2022potato}. The annotation interface setup is available at {\small \url{https://github.com/davidjurgens/potato/tree/master/example-projects/match_finding}} which includes all the following instructions as well.

\paragraph{Task description:} The task is to label to what degree two sentences have the same information. The information in the sentences is scientific findings. Here, a scientific finding is a statement that describes a research output of a scientific study, such as a result, conclusion, product, etc. You should rate how similar the findings are; you can ignore extra information like ``The researchers showed...'', ``In vivo experiments demonstrated...'' etc. For example, in the sentence ``After controlling for weight and age, researchers found that overconsumption of sugar is linked with an increase in diabetes,'' the information in the finding is ``overconsumption of sugar is linked with an increase in diabetes''. Some sentences may have no findings or multiple findings, so use your best judgment about what are the core findings being said.

You will rate this on a 5-point scale, where each level means the following:

\begin{enumerate}
\item The information in the findings is completely different
\begin{itemize}
    \item Sentences in this category have findings which say completely different information
    \item The sentences may be on totally different topics
    \begin{itemize}
        \item Overconsumption of sugar causes diabetes
        \item Regular exercise improves heart health
    \end{itemize}
    \item There may be some overlap in key words used between the two sentences, but the actual information is completely different
    \begin{itemize}
        \item Chocolate contains a lot of sugar, and therefore can have an effect on weight.
        \item Overconsumption of sugar leads to diabetes.
    \end{itemize}
\end{itemize}

\item The information in the findings is mostly different
\begin{itemize}
    \item The findings may talk about the same topic, but the actual information is mostly different; for example, these sentences convey mostly different information even though they talk about the same topic:
    \begin{itemize}
        \item Overconsumption of sugar causes diabetes
        \item Sugar is good for your health
    \end{itemize}
    \item There could be a link between the two findings, but the information conveyed is still different
    \begin{itemize}
        \item Overconsumption of sugar increases blood glucose levels
        \item High blood glucose over time increases the risk of developing diabetes
    \end{itemize}
\end{itemize}

\item The information in the findings is somewhat similar
\begin{itemize}
    \item The findings are discussing relevant research outputs but there are some differences in the information conveyed. Here the difference is that (i) talks about the relationship between overconsumption of sugar and diabetes and (ii) describes how genetics plays a role in overconsumption of sugar
    \begin{itemize}
        \item Overconsumption of sugar causes diabetes
        \item Overconsumption of sugar might be genetically determined
    \end{itemize}
\end{itemize}

\item The information in the findings is mostly the same
\begin{itemize}
    \item In this case there may be some changes in e.g. the level of generality. Additionally, one sentence may go into more detail than the other and add additional context, but the information is largely the same
    \item Here the two findings have the same information but at different levels of generality:
    \begin{itemize}
        \item A link between sugar and diabetes was found
        \item Overconsumption of sugar is associated with the onset of diabetes
    \end{itemize}
    \item Here both sentences have the same core finding, but one sentence goes into more detail
    \begin{itemize}
        \item Overconsumption of sugar causes diabetes
        \item Experiments demonstrated that overconsumption of sugar led to an increase in blood glucose levels, which over a long enough time period was linked to an increased prevalence of diabetes in the cohort.
    \end{itemize}
    \item One finding could support the other
    \begin{itemize}
        \item Overconsumption of sugar causes diabetes
        \item Overconsumption of sugar can have negative effects on health
    \end{itemize}
\end{itemize}

\item The information in the findings is completely the same
\begin{itemize}
    \item In this case there is complete overlap in the information in the findings conveyed by the two sentences
    \begin{itemize}
        \item Overconsumption of sugar leads to diabetes.
        \item The researchers found that overconsumption of sugar leads to diabetes
    \end{itemize}
    \item Note that there can be changes in e.g. the level of certainty or the strength of the information.
    \begin{itemize}
        \item Overconsumption of sugar leads to diabetes.
        \item It is likely that there is a link between overconsumption of sugar and the onset of diabetes.

    \end{itemize}
\end{itemize}
\end{enumerate}

\section{Final dataset details}
\label{sec:dataset-details-supplement}
\begin{figure}[h]
  
  \centering
    \includegraphics[width=\linewidth]{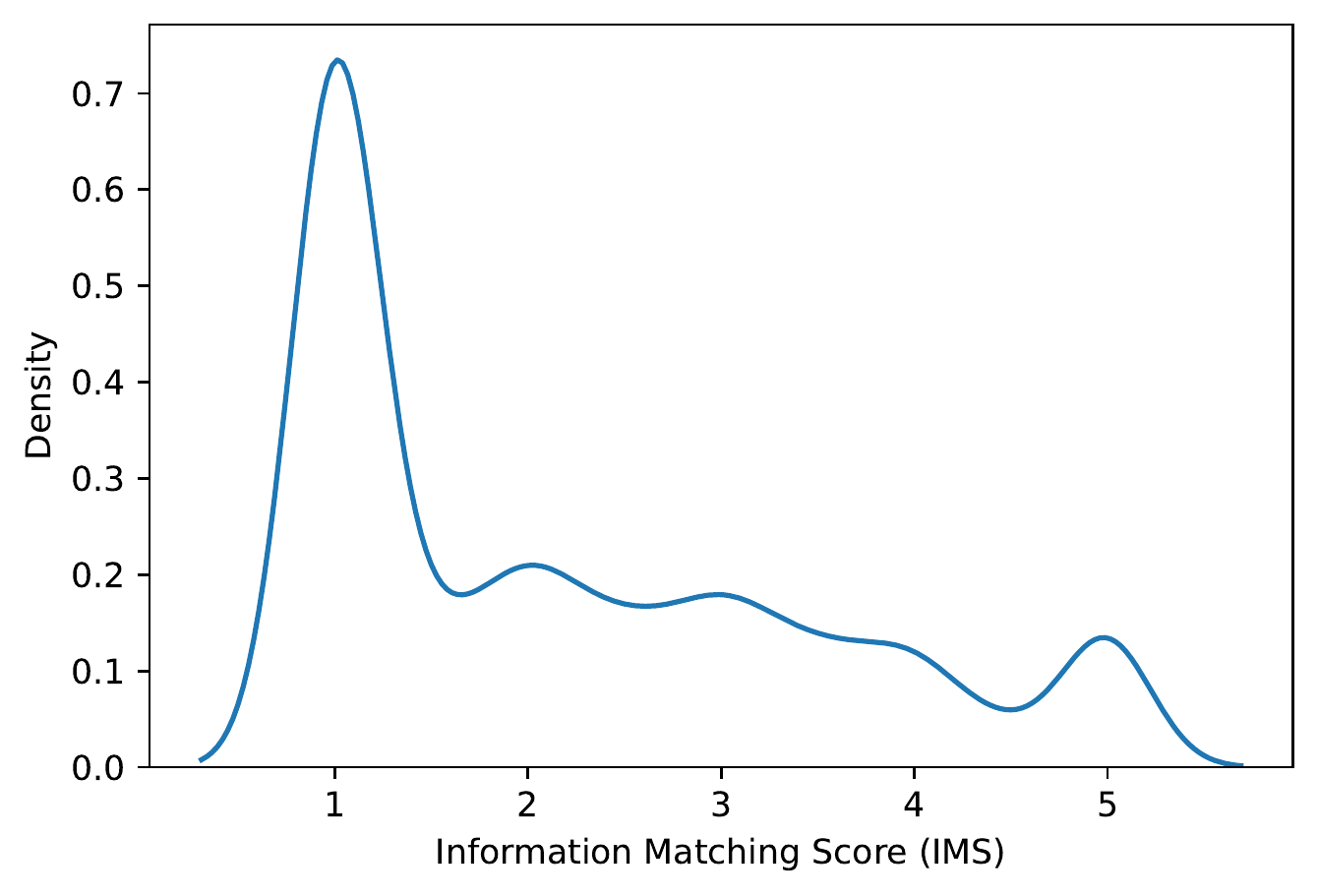}
    \caption{Distribution of the final matching score in \DATASET{}, which includes some pairs of scientific findings that are automatically labeled based on their extreme textual similarity (high or low), in addition to the annotated pairs.
    }
    \label{fig:matching_score_distribution}
\end{figure}

\begin{figure}[h]
  
  \centering
    \includegraphics[width=\linewidth]{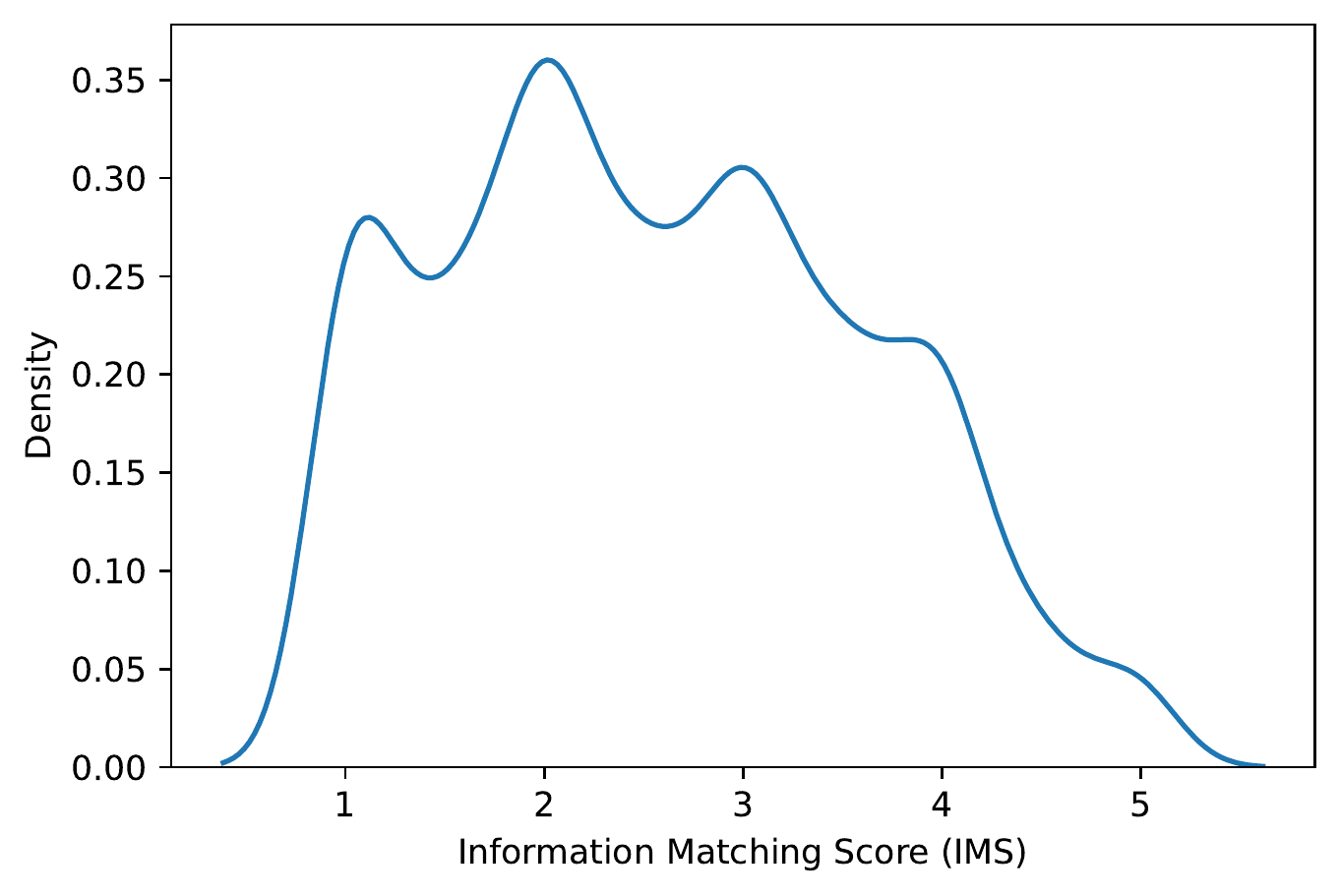}
    \caption{Distribution of the final matching score for annotated pairs in \DATASET{}
    }
    \label{fig:matching_score_distribution_no_easy}
\end{figure}

\begin{figure*}[t]
  
  \centering
    \includegraphics[width=\linewidth]{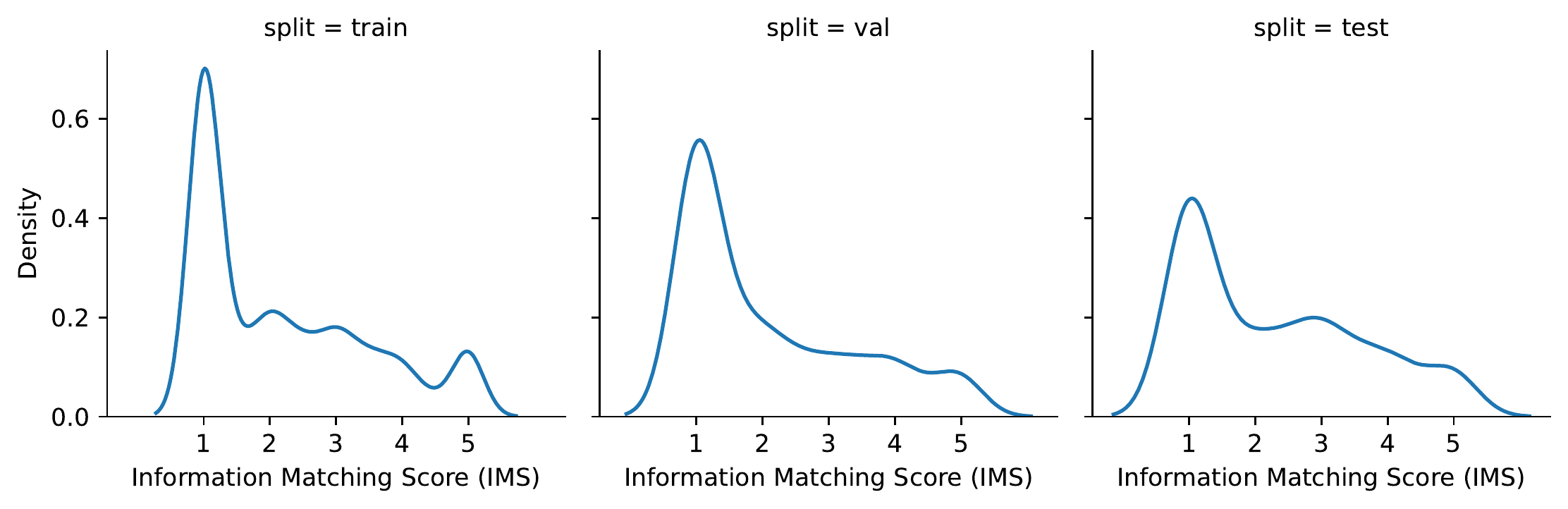}
    \caption{Distribution of the final matching score for each split set in \DATASET{}
    }
    \label{fig:matching_score_distribution_splits}
\end{figure*}

Figure \ref{fig:matching_score_distribution} shows the IMS distribution in \DATASET{}. Figure \ref{fig:matching_score_distribution_no_easy} shows the IMS distribution for annotated pairs in \DATASET{}. Figure \ref{fig:matching_score_distribution_splits} shows the IMS distribution for each split.

We measure various aspects of lexical richness between the different domains of the data in \autoref{tab:lexical-richness}.

\begin{table}
    \setlength{\tabcolsep}{1.5pt}
    \def\arraystretch{1.2}
    \centering
    \fontsize{10}{10}\selectfont
    \begin{tabular}{l c c c c}
    \toprule 
    Metric & Papers & Overall & News & Tweets\\
    \midrule 

        Unique tokens & $11047$ & $12139$ & $10203$ & $5037$\\
        RTTR & $32.01$ & $36.59$ & $33.48$ & $38.46$\\
        MTLD & $152.64$ & $185.35$ & $176.53$ & $259.88$\\
        HDD & $0.89$ & $0.90$ & $0.89$ & $0.92$\\
    
    \bottomrule 

    \end{tabular}
    \caption{Various measures of lexical richness and diversity between findings in papers and other sources. RTTR is the root token-type ratio; MTLD is measure of textual lexical diversity~\cite{mccarthy2010mtld}; HDD is the hypergeometric distribution diversity~\cite{mccarthy2010mtld}.}
    \label{tab:lexical-richness}
\end{table}

\section{Metrics\footnote{We use relevant libraries for all metrics e.g. \texttt{sklearn.metrics}}}
\label{sec:metrics-supplement}
\paragraph{Average Normalized Edit Distance} We calculate the normalized edit distance as follows:
\begin{equation*}
d_{N} = \frac{1}{|D|}\sum_{i}\frac{d(s_{1}^{(i)},s_{2}^{(i)})}{\max{(|s_{1}^{(i)}|,|s_{2}^{(i)}|)}}
\end{equation*}
where $|D|$ is the size of the dataset, $(s_{1}^{(i)},s_{2}^{(i)})$ is a sentence pair, and $d$ is the edit distance. 

\paragraph{Jaccard Index} The Jaccard index is calculated based on the overlap of the members of two sets (e.g. the words in two sentences $X$ and $Y$):
\begin{equation*}
    J = \frac{|X \cap Y|}{|X \cup Y|}
\end{equation*}


\paragraph{Cosine Similarity} The cosine similarity between two vectors $\textbf{a}$ and $\textbf{b}$ is calculated as:
\begin{equation*}
    S_{C}(\textbf{a},\textbf{b)} = \frac{\textbf{a} \cdot \textbf{b}}{\lVert\textbf{a}\rVert \lVert\textbf{b}\rVert}
\end{equation*}
Which is their dot product divided by the product of their lengths.

\paragraph{Mean Squared Error} The mean squared error between two lists of numbers of length $n$ is calculated as:
\begin{equation*}
    \text{MSE}(Y, \hat{Y}) = \frac{1}{n}\sum_{i}(Y_{i} - \hat{Y}_{i})^{2}
\end{equation*}

\paragraph{Mean Average Precision} The mean average precision in ranking takes the average Precision@k (P@k) for every relevant sample in a ranked list. First, P@k is calculated as follows:
\begin{equation*}
    \text{P@k}(\hat{Y}) = \frac{1}{k}\sum_{i}^{k}\mathbbm{1}(\hat{Y}_{i} = 1)
\end{equation*}
where $\mathbbm{1}$ is the indicator function. The average precision is then taken over all relevant items in the list, where there are $r$ relevant items:
\begin{equation*}
    \text{AP}(\hat{Y}) = \frac{1}{r}\sum_{k}\text{P@k}(\hat{Y}[:k]) \text{ where } \hat{Y}_k = 1
\end{equation*}
The mean average precision for a set of $n$ ranked lists $D$ is then the mean of the average precision of each of these lists:
\begin{equation*}
    \text{MAP} = \frac{1}{n}\sum_{j}\text{AP}(D_{n})
\end{equation*}

\paragraph{Mean Reciprocal Rank} The mean reciprocal rank (MRR) calculates the mean rank for each relevant item in a list i.e. its position in that list. It is calculated as follows for $D$ lists or relevant items in $\hat{Y}$ ranked lists:
\begin{equation*}
    \text{MRR}(D) = \frac{1}{|D|}\sum_{j}\frac{1}{|D_{j}|}\sum_{i}\frac{1}{\text{rank}_i(\hat{Y}_{j})}
\end{equation*}
where $\text{rank}_i(\hat{Y}_{j})$ is the rank of item $i$ in list $\hat{Y}_{j}$.

\section{Full Model Details}
\label{sec:full-model-descriptions}

All baseline experiments were run on a shared cluster. Requested jobs consisted of 16GB of RAM and 4 Intel Xeon Silver 4110 CPUs. We used a single NVIDIA Titan RTX GPU for experiments. Training takes approximately 3 minutes for all MLM-based models and 2 minutes for SBERT models. 

\paragraph{RoBERTa}
RoBERTa is a large pretrained transformer language model, trained using the masked language modeling (MLM) objective on a large corpus of English text. We use the base model of RoBERTa for our experiments. Huggingface model name: roberta-base -- 124,647,170 parameters

\paragraph{MiniLM} We use a popular pretrained SBERT model based on MiniLM~\cite{wang2020minilm}, which is trained by distilling multiple language models into one compressed model. SBERT uses siamese BERT encoders to obtain sentence embeddings for pairs of sentences and is trained to decrease the distance between these two embeddings. The pretraining for the sentence similarity task consists of a wide range of datasets covering multiple domains and $>$ 1 billion sentence pairs, including science~\cite{DBLP:conf/acl/CohanFBDW20, DBLP:conf/acl/LoWNKW20}. As much of the data is collected automatically, it uses a contrastive learning objective where known relevant pairs are treated as positive values and other samples in a batch are treated as negative values. The model is then trained to minimize the cross-entropy between the dot-product of embeddings and the label acquired from positive/negative samples.
Huggingface model name (sentence transformers): all-MiniLM-L6-v2 -- 22,713,216 parameters

\paragraph{MPNet} This is the same setup as in MiniLM but with using MPNet as the base network~\cite{DBLP:conf/nips/Song0QLL20}. MPNet is trained using a permuted language modeling (PLM) objective with position information as input to achieve the best of both worlds between MLM and PLM. The base network is used in the SBERT setup where it is further fine-tuned on the same dataset and same task as with MiniLM

Huggingface model name (sentence transformers): all-mpnet-base-v2 -- 109,486,464 parameters

\paragraph{Paraphrase Detection}
This is a paraphrase detection model based on RoBERTa used in~\cite{nighojkar-licato-2021-improving}. The model is trained on the adversarial paraphrase dataset introduced in that paper. 

Huggingface model name (sentence transformers): coderpotter/adversarial-paraphrasing-detector -- 124,647,170 parameters

\paragraph{NLI}
This is a RoBERTa model trained on a wide array of NLI datasets, including SNLI~\cite{snli:emnlp2015}, MNLI~\cite{N18-1101}, FEVER (a fact-checking dataset)~\cite{DBLP:conf/naacl/ThorneVCM18} and ANLI~\cite{nie-etal-2020-adversarial}.

Huggingface model name (sentence transformers): ynie/roberta-large-snli\_mnli\_fever\_anli\_R1\_R2\_R3-nli -- 124,647,170 parameters

\paragraph{SciBERT}
SciBERT is the original BERT model trained using MLM on a large set of scientific papers from Semantic Scholar~\cite{DBLP:conf/acl/LoWNKW20}. 

Huggingface model name (sentence transformers): allenai/scibert\_scivocab\_uncased -- 109,920,514 parameters

\paragraph{CiteBERT}
CiteBERT is SciBERT further fine-tuned on the CiteWorth dataset for the task of citation detection, which predicts if a given sentence requires a citation or not~\cite{wright2021citeworth}.

Huggingface model name (sentence transformers): copenlu/citebert -- 109,920,514 parameters

We use sane defaults when fine-tuning each of our models. In this, for the MLM based models we use [lr: 2e-5, n\_epochs: 3, warmup\_steps: 200, weight\_decay: 0.01, batch\_size: 8]. For SBERT models we use the same setting except we train for 5 epochs.

\section{Exaggeration Detection}
\label{sec:exaggeration-supplement}
The problem of scientific exaggeration detection was studied in~\cite{wright2021semi}. The basic task is: given a pair of scientific findings (e.g. a reference finding from a paper and its counterpart in a news article), determine if one finding is exaggerating the other finding. More formally, the task focuses on differences in the causal claim strength of the two findings, where the claim strength can take on one of four values:
\begin{itemize}[noitemsep,nolistsep]
    \item 0: No statement of relationship
    \item 1: Correlational statement (e.g. ``X is associated with Y'')
    \item Conditional causal statement (e.g. ``X might cause Y under circumstance Z'')
    \item Causal statement (e.g. ``X causes Y'')
\end{itemize}
\citet{wright2021semi} curate data and build models for performing the exaggeration detection task in two different settings: as predicting the individual claim strengths and comparing, and as an inference task where a model is fed both findings and asked to predict if the reference finding is being exaggerated, downplayed, or faithfully represented by its counterpart. We use the best-performing model from their paper, which is a multi-task few-shot learning model based on pattern exploiting training (PET) called MT-PET. In particular, we use the model for strength classification which has seen 4,500 individual findings labeled for claim strength and 200 pairs labeled for exaggeration.

\section{Scientific Text Parser}
\label{sec:parser-supplement}
We fine-tuned a RoBERTa model over 200K self-labeled abstracts from PubMed. The model is trained to predict five labels including: BACKGROUND, CONCLUSIONS, METHODS, OBJECTIVE and RESULTS. We did a 8:1:1 split for the data and fine-tune the RoBERTa model for 1 epoch. 0.92 F1 is attained on the test set.

\section{Extended Benchmarking}
\label{sec:extended-benchmarking}
Tables with extended benchmarking results can be found in \autoref{tab:baseline-findings-matching} to \autoref{tab:cs-baseline-findings-matching}.

\begin{table*}
    \setlength{\tabcolsep}{1.5pt}
    \def\arraystretch{1.2}
    \centering
    \fontsize{10}{10}\selectfont
    \rowcolors{2}{gray!10}{white}
    \begin{tabular}{l c c | c c | c c}
    \toprule 
    & \multicolumn{2}{c}{All} & \multicolumn{2}{c}{News} & \multicolumn{2}{c}{Twitter} \\
     \midrule
    Method & MSE & $\rho$ & MSE & $\rho$ & MSE & $\rho$ \\
    \midrule 
Paraphrase & $3.170_{0.000}$ & $23.58_{0.00}$ & $3.310_{0.000}$ & $19.24_{0.00}$ & $2.824_{0.000}$ & $35.41_{0.00}$ \\
NLI & $2.921_{0.000}$ & $35.71_{0.00}$ & $2.786_{0.000}$ & $35.78_{0.00}$ & $3.255_{0.000}$ & $34.76_{0.00}$ \\
MiniLM & $0.628_{0.000}$ & $73.98_{0.00}$ & $0.646_{0.000}$ & $76.27_{0.00}$ & $0.583_{0.000}$ & $64.61_{0.00}$ \\
MPNet & $0.718_{0.000}$ & $72.59_{0.00}$ & $0.713_{0.000}$ & $74.76_{0.00}$ & $0.730_{0.000}$ & $62.06_{0.00}$ \\
\midrule
SciBERT & $0.579_{0.011}$ & $73.24_{0.73}$ &  $0.596_{0.018}$ & $74.66_{0.75}$ &  $0.538_{0.021}$ & $66.29_{0.67}$\\
CiteBERT & $0.581_{0.027}$ & $73.37_{0.78}$ &  $0.592_{0.034}$ & $74.81_{0.91}$ &  $0.552_{0.030}$ & $66.13_{1.43}$\\
RoBERTa & $0.587_{0.017}$ & $74.44_{0.81}$ &  $0.602_{0.033}$ & $75.82_{0.71}$ &  $0.550_{0.067}$ & $\mathbf{68.66_{1.29}}$\\
MiniLM-FT & $0.492_{0.001}$ & $75.84_{0.03}$ &  $\mathbf{0.465_{0.001}}$ & $78.66_{0.05}$ &  $0.559_{0.002}$ & $63.80_{0.05}$\\
MPNet-FT & $\mathbf{0.489_{0.003}}$ & $\mathbf{76.48_{0.07}}$ &  $0.474_{0.003}$ & $\mathbf{78.71_{0.17}}$ &  $\mathbf{0.526_{0.008}}$ & $66.45_{0.37}$\\

    \bottomrule 

    \end{tabular}
    \caption{OVERALL --MSE and Pearson correlation ($\rho$) on predicting the similarity of the scientific findings for different models. Results are averaged over 5 random seeds; standard deviation is given in the subscript.} 
    \label{tab:baseline-findings-matching}
\end{table*}

\begin{table*}
    \setlength{\tabcolsep}{1.5pt}
    \def\arraystretch{1.2}
    \centering
    \fontsize{10}{10}\selectfont
    \rowcolors{2}{gray!10}{white}
    \begin{tabular}{l c c | c c | c c}
    \toprule 
    & \multicolumn{2}{c}{All} & \multicolumn{2}{c}{News} & \multicolumn{2}{c}{Twitter} \\
     \midrule
    Method & MSE & $\rho$ & MSE & $\rho$ & MSE & $\rho$ \\
    \midrule 
Paraphrase & $2.773_{0.000}$ & $27.16_{0.00}$ & $2.846_{0.000}$ & $30.22_{0.00}$ & $2.577_{0.000}$ & $28.18_{0.00}$ \\
NLI & $2.529_{0.000}$ & $40.23_{0.00}$ & $2.225_{0.000}$ & $47.55_{0.00}$ & $3.339_{0.000}$ & $6.23_{0.00}$ \\
MiniLM & $0.618_{0.000}$ & $76.45_{0.00}$ & $0.658_{0.000}$ & $80.31_{0.00}$ & $\mathbf{0.509_{0.000}}$ & $\mathbf{63.78_{0.00}}$ \\
MPNet & $0.804_{0.000}$ & $73.14_{0.00}$ & $0.815_{0.000}$ & $76.91_{0.00}$ & $0.777_{0.000}$ & $56.11_{0.00}$ \\
\midrule
SciBERT & $0.554_{0.020}$ & $71.67_{0.94}$ &  $0.507_{0.026}$ & $76.69_{0.73}$ &  $0.681_{0.058}$ & $43.56_{5.32}$\\
CiteBERT & $0.542_{0.031}$ & $72.55_{0.92}$ &  $0.496_{0.034}$ & $77.31_{1.11}$ &  $0.663_{0.029}$ & $46.01_{2.61}$\\
RoBERTa & $0.511_{0.036}$ & $75.40_{1.19}$ &  $0.475_{0.035}$ & $79.33_{0.78}$ &  $0.608_{0.056}$ & $53.72_{4.56}$\\
MiniLM-FT & $\mathbf{0.377_{0.002}}$ & $\mathbf{79.46_{0.15}}$ &  $\mathbf{0.327_{0.003}}$ & $\mathbf{84.08_{0.14}}$ &  $0.512_{0.001}$ & $60.00_{0.23}$\\
MPNet-FT & $0.412_{0.005}$ & $77.98_{0.23}$ &  $0.361_{0.004}$ & $82.30_{0.22}$ &  $0.548_{0.013}$ & $57.79_{0.72}$\\

    \bottomrule 

    \end{tabular}
    \caption{BIOLOGY -- MSE and Pearson correlation ($\rho$) on predicting the similarity of the scientific findings for different models. Results are averaged over 5 random seeds; standard deviation is given in the subscript.} 
    \label{tab:bio-baseline-findings-matching}
\end{table*}

\begin{table*}
    \setlength{\tabcolsep}{1.5pt}
    \def\arraystretch{1.2}
    \centering
    \fontsize{10}{10}\selectfont
    \rowcolors{2}{gray!10}{white}
    \begin{tabular}{l c c | c c | c c}
    \toprule 
    & \multicolumn{2}{c}{All} & \multicolumn{2}{c}{News} & \multicolumn{2}{c}{Twitter} \\
     \midrule
    Method & MSE & $\rho$ & MSE & $\rho$ & MSE & $\rho$ \\
    \midrule 
Paraphrase & $3.282_{0.000}$ & $15.95_{0.00}$ & $3.525_{0.000}$ & $31.32_{0.00}$ & $2.629_{0.000}$ & $29.56_{0.00}$ \\
NLI & $2.820_{0.000}$ & $37.03_{0.00}$ & $2.841_{0.000}$ & $34.60_{0.00}$ & $2.763_{0.000}$ & $49.39_{0.00}$ \\
MiniLM & $0.706_{0.000}$ & $76.46_{0.00}$ & $0.739_{0.000}$ & $78.34_{0.00}$ & $0.615_{0.000}$ & $62.92_{0.00}$ \\
MPNet & $0.738_{0.000}$ & $79.41_{0.00}$ & $0.726_{0.000}$ & $81.42_{0.00}$ & $0.771_{0.000}$ & $64.96_{0.00}$ \\
\midrule
SciBERT & $0.429_{0.039}$ & $81.44_{1.44}$ &  $0.440_{0.027}$ & $83.37_{1.31}$ &  $0.400_{0.085}$ & $\mathbf{70.35_{2.90}}$\\
CiteBERT & $0.431_{0.044}$ & $81.80_{1.19}$ &  $0.433_{0.042}$ & $83.92_{1.32}$ &  $0.425_{0.067}$ & $69.49_{1.21}$\\
RoBERTa & $0.437_{0.040}$ & $82.20_{0.60}$ &  $0.425_{0.046}$ & $84.77_{1.12}$ &  $0.470_{0.185}$ & $69.73_{5.02}$\\
MiniLM-FT & $0.436_{0.004}$ & $79.31_{0.15}$ &  $0.445_{0.003}$ & $81.80_{0.11}$ &  $0.412_{0.007}$ & $64.08_{0.47}$\\
MPNet-FT & $\mathbf{0.371_{0.005}}$ & $\mathbf{82.58_{0.17}}$ &  $\mathbf{0.369_{0.005}}$ & $\mathbf{85.20_{0.22}}$ &  $\mathbf{0.377_{0.008}}$ & $65.03_{0.38}$\\

    \bottomrule 

    \end{tabular}
    \caption{MEDICINE -- MSE and Pearson correlation ($\rho$) on predicting the similarity of the scientific findings for different models. Results are averaged over 5 random seeds; standard deviation is given in the subscript.} 
    \label{tab:med-baseline-findings-matching}
\end{table*}

\begin{table*}
    \setlength{\tabcolsep}{1.5pt}
    \def\arraystretch{1.2}
    \centering
    \fontsize{10}{10}\selectfont
    \rowcolors{2}{gray!10}{white}
    \begin{tabular}{l c c | c c | c c}
    \toprule 
    & \multicolumn{2}{c}{All} & \multicolumn{2}{c}{News} & \multicolumn{2}{c}{Twitter} \\
     \midrule
    Method & MSE & $\rho$ & MSE & $\rho$ & MSE & $\rho$ \\
    \midrule 
Paraphrase & $3.208_{0.000}$ & $32.23_{0.00}$ & $3.568_{0.000}$ & $27.56_{0.00}$ & $2.618_{0.000}$ & $46.52_{0.00}$ \\
NLI & $3.066_{0.000}$ & $39.57_{0.00}$ & $3.125_{0.000}$ & $27.39_{0.00}$ & $2.970_{0.000}$ & $50.61_{0.00}$ \\
MiniLM & $0.539_{0.000}$ & $75.16_{0.00}$ & $0.525_{0.000}$ & $77.98_{0.00}$ & $0.561_{0.000}$ & $66.81_{0.00}$ \\
MPNet & $0.634_{0.000}$ & $72.22_{0.00}$ & $0.650_{0.000}$ & $72.26_{0.00}$ & $0.608_{0.000}$ & $69.44_{0.00}$ \\
\midrule
SciBERT & $0.531_{0.022}$ & $74.57_{1.36}$ &  $0.571_{0.020}$ & $74.68_{1.56}$ &  $\mathbf{0.467_{0.030}}$ & $\mathbf{74.14_{1.41}}$\\
CiteBERT & $0.555_{0.015}$ & $73.23_{0.39}$ &  $0.585_{0.036}$ & $73.68_{0.52}$ &  $0.505_{0.031}$ & $72.50_{1.29}$\\
RoBERTa & $0.655_{0.040}$ & $71.28_{1.24}$ &  $0.720_{0.085}$ & $71.38_{1.88}$ &  $0.550_{0.057}$ & $71.35_{1.58}$\\
MiniLM-FT & $\mathbf{0.500_{0.004}}$ & $\mathbf{75.52_{0.11}}$ &  $\mathbf{0.467_{0.005}}$ & $\mathbf{78.48_{0.12}}$ &  $0.555_{0.004}$ & $66.50_{0.16}$\\
MPNet-FT & $0.520_{0.009}$ & $75.21_{0.25}$ &  $0.550_{0.006}$ & $75.48_{0.18}$ &  $0.471_{0.014}$ & $72.25_{0.67}$\\

    \bottomrule 

    \end{tabular}
    \caption{PSYCHOLOGY -- MSE and Pearson correlation ($\rho$) on predicting the similarity of the scientific findings for different models. Results are averaged over 5 random seeds; standard deviation is given in the subscript.} 
    \label{tab:psych-baseline-findings-matching}
\end{table*}

\begin{table*}
    \setlength{\tabcolsep}{1.5pt}
    \def\arraystretch{1.2}
    \centering
    \fontsize{10}{10}\selectfont
    \rowcolors{2}{gray!10}{white}
    \begin{tabular}{l c c | c c | c c}
    \toprule 
    & \multicolumn{2}{c}{All} & \multicolumn{2}{c}{News} & \multicolumn{2}{c}{Twitter} \\
     \midrule
    Method & MSE & $\rho$ & MSE & $\rho$ & MSE & $\rho$ \\
    \midrule 
Paraphrase & $3.373_{0.000}$ & $24.35_{0.00}$ & $3.346_{0.000}$ & $26.48_{0.00}$ & $3.463_{0.000}$ & $37.48_{0.00}$ \\
NLI & $3.177_{0.000}$ & $29.97_{0.00}$ & $2.945_{0.000}$ & $36.51_{0.00}$ & $3.926_{0.000}$ & $-8.74_{0.00}$ \\
MiniLM & $0.656_{0.000}$ & $71.40_{0.00}$ & $0.656_{0.000}$ & $73.09_{0.00}$ & $0.656_{0.000}$ & $66.64_{0.00}$ \\
MPNet & $0.705_{0.000}$ & $70.03_{0.00}$ & $0.670_{0.000}$ & $72.43_{0.00}$ & $0.815_{0.000}$ & $60.18_{0.00}$ \\
\midrule
SciBERT & $0.738_{0.020}$ & $67.66_{0.71}$ &  $0.777_{0.031}$ & $67.46_{1.03}$ &  $0.609_{0.029}$ & $69.97_{1.76}$\\
CiteBERT & $0.733_{0.045}$ & $68.05_{1.38}$ &  $0.770_{0.051}$ & $67.83_{1.34}$ &  $0.612_{0.040}$ & $69.79_{2.33}$\\
RoBERTa & $0.690_{0.021}$ & $71.53_{1.15}$ &  $0.731_{0.031}$ & $71.24_{0.80}$ &  $\mathbf{0.560_{0.075}}$ & $\mathbf{75.49_{3.09}}$\\
MiniLM-FT & $0.611_{0.003}$ & $72.32_{0.05}$ &  $0.577_{0.001}$ & $74.13_{0.06}$ &  $0.721_{0.008}$ & $66.44_{0.25}$\\
MPNet-FT & $\mathbf{0.603_{0.004}}$ & $\mathbf{73.00_{0.20}}$ &  $\mathbf{0.575_{0.006}}$ & $\mathbf{74.46_{0.36}}$ &  $0.692_{0.011}$ & $67.18_{0.59}$\\

    \bottomrule 

    \end{tabular}
    \caption{COMPUTER SCIENCE -- MSE and Pearson correlation ($\rho$) on predicting the similarity of the scientific findings for different models. Results are averaged over 5 random seeds; standard deviation is given in the subscript.} 
    \label{tab:cs-baseline-findings-matching}
\end{table*}

\section{Error Examples}
\label{sec:error-examples}
Examples of errors which our best models made on $\langle$tweet, paper$\rangle$ pairs can be found in \autoref{tab:roberta-tweet-errors} and \autoref{tab:mpnet-tweet-errors}.

\begin{table*}[t]
\small

\newcommand{\tabincell}[2]{\begin{tabular}{@{}#1@{}}#2\end{tabular}}
\resizebox{0.99\textwidth}{!}{
\begin{tabular}{p{60mm}p{60mm}cc}
\toprule
\textbf{Tweet} & \textbf{Paper Finding} & \textbf{Prediction} & \textbf{Ground Truth} \\
\midrule
Mixed reality variations improve learning, over screen-only options. CMU researchers.  & The overall improvement from pre to post was 11.3 \% in the mixed-reality conditions and 2.4 \% in the virtual conditions. & 2.92 & 5 \\ \midrule
Metarrestin, an inhibitor of tumor metastasis, discovered thru team science @ KU, @username, @username, and @username and more. Congrats to first author Kevin Frankowski and special thanks to Udo Rudloff, Juan Maruguan, and Sui Huang. & Evaluation of apoptotic index showed less than 1\% of cells undergoing apoptosis in response to metarrestin treatment. & 2.15 & 4.2 \\
\midrule
Today in @username a graphene transfer approach using paraffin as a support layer to obtain wrinkle-reduced, clean, large-area graphene retaining high mobility & Similar to previous reports, our PMMA-transferred CVD monolayer graphene on Si/SiO 2 substrate experienced compressive strain and p-doping 30 . & 2.64 & 1 \\
\midrule
When the Going Gets Tough: The "Why” of Goal Striving Matters. An excellent article by @username + colleagues. & Practitioners who aim to facilitate effective goal setting in sport, business, and educational settings would benefit from guidelines for developing autonomous motivation. & 2.00 & 3.6 \\
\midrule
Thosewho were sociosexually unrestricted reported lower stress and greater overall emotional health after casual sex. & Simple slope analyses indicated that high-SOI participants who had casual sex over the academic year had higher self-esteem (B $\frac{1}{4}$ 0.14, SE $\frac{1}{4}$ 0.06, p $\frac{1}{4}$ .025) and marginally lower depression (B $\frac{1}{4}$ A0.12, SE $\frac{1}{4}$ 0.07, p $\frac{1}{4}$ .091) and anxiety (B $\frac{1}{4}$ A0.11, SE $\frac{1}{4}$ 0.06, p $\frac{1}{4}$ .086) than high-SOI participants who did not have casual sex (Figure 3) . & 2.84 & 4.4\\
\bottomrule
\end{tabular}
}
\caption{Top-5 biggest errors made by RoBERTa on <tweet, paper> pairs in terms of absolute error. }
\label{tab:roberta-tweet-errors}
\end{table*}

\begin{table*}[t]
\small

\newcommand{\tabincell}[2]{\begin{tabular}{@{}#1@{}}#2\end{tabular}}
\resizebox{0.99\textwidth}{!}{
\begin{tabular}{p{60mm}p{60mm}cc}
\toprule
\textbf{Tweet} & \textbf{Paper Finding} & \textbf{Prediction} & \textbf{Ground Truth} \\
\midrule
Mixed reality variations improve learning, over screen-only options. CMU researchers.  & The overall improvement from pre to post was 11.3 \% in the mixed-reality conditions and 2.4 \% in the virtual conditions. & 2.45 & 5 \\
\midrule
'Physical observation + interactive feedback improved children’s learning by 5x' via Nesra Yannier @username & These results show that mixed-reality led to more learning than screen only, for both the mousecontrol and physical-control conditions ( Figure 10 ). & 2.19 & 4 \\
\midrule
Today in @username a graphene transfer approach using paraffin as a support layer to obtain wrinkle-reduced, clean, large-area graphene retaining high mobility & Similar to previous reports, our PMMA-transferred CVD monolayer graphene on Si/SiO 2 substrate experienced compressive strain and p-doping 30 . & 2.80 & 1 \\
\midrule
Metarrestin, an inhibitor of tumor metastasis, discovered thru team science @ KU, @username, @username, and @username and more. Congrats to first author Kevin Frankowski and special thanks to Udo Rudloff, Juan Maruguan, and Sui Huang. & Evaluation of apoptotic index showed less than 1\% of cells undergoing apoptosis in response to metarrestin treatment. & 2.61 & 4.2 \\
\midrule
Super happy to present our latest paper on global food webs:
Years of work on predator-prey body-mass ratios and the first use of the GATEWAy data base. & Predators typically exert the strongest feeding pressure on prey that are 1-2 orders of magnitude smalle, while weaker interaction strengths are realized with prey that are smaller or larger than this size. & 1.92 & 3.4\\
\bottomrule
\end{tabular}
}
\caption{Top-5 biggest errors made by MPNet-FT on <tweet, paper> pairs in terms of absolute error. }
\label{tab:mpnet-tweet-errors}
\end{table*}

\begin{table*}[t]
\small

\newcommand{\tabincell}[2]{\begin{tabular}{@{}#1@{}}#2\end{tabular}}
\resizebox{0.99\textwidth}{!}{
\rowcolors{2}{gray!10}{white}
\begin{tabular}{p{60mm}p{60mm}cc}
\toprule
\textbf{Paper Finding} & \textbf{News Finding} & \textbf{Prediction} \\
\midrule
Increase in the body size of dicynodonts across the Late Triassic may have been driven by selection pressure to reach a size refuge from large predators (24) .  & Researchers believe selection pressures--potentially to protect themselves from larger predators--may have been the driver behind their giant size, but more research will be needed to understand Lisowicia and its place in the evolutionary tree. & 3.0008 \\ \midrule
The best option among the three is the EPS container with the lowest impacts across the 12 categories. & The study found that the styrofoam container was the best option among the disposable containers across all the impacts considered, including the carbon footprint. & 3.1120 \\
\midrule
As media coverage started to increase, water demand decreased and the models with media correctly captured the downward trend, but the models without media forecasted increasing demand. & Strikingly, the models also found that for every 100-article increase over a two-month period, there was an 11 percent to 18 percent decrease in demand for water. & 3.1537 \\
\midrule
For example, of the 63 negative precipitation years during 1896-2014, 15 of the 32 warm-dry years (47\%) produced 1-SD drought, compared with only 5 of the 31 cool-dry years (16\%)  & Their analysis revealed that the years that were both warm and dry were about twice as likely to produce a severe drought as years that were cool and dry. & 3.2569 \\
\midrule
Our study shows that low-dose BPA and BPS exposure has physiological effects. & Although the levels were low, the scientists soon saw that both BPA and BPS caused changes in the brain development of the zebra fish embryos. & 3.3331 \\
\midrule
Use of multiple prescription medications with these potential effects was associated with greater likelihood of concurrent depression. & About 15 percent of participants who simultaneously used three or more of these drugs were depressed. & 3.3692 \\ 
\midrule
We also found that renewal submission rate was the factor most predictive of sustained funding for either gender, and that gender differences in survival disappear when genders were matched on renewal submission rate and first year of funding. & On average, women submitted eligible grants for renewal 42\% of the time and won funding 36\% of the time, compared with 45\% and 39\%, respectively, for men. & 3.4132 \\
\midrule
Among those completing the 12-month survey, 60 nonsmokers (55.6\%) and 29 smokers (26.6\%) were reemployed at 1 year. & After 12 months, the re-employment rate of smokers was 24 percent lower than that of nonsmokers. & 3.5151 \\
\midrule
This suggests behaviour consistent with moral licensing: participants who refrained from cheating at higher stakes seem to have subsequently licensed themselves to donate less to charity, thereby "balancing" their moral behaviour over time. & However those who cheated the least when tempted with high stakes were more likely to license themselves not to behave so charitably in another task. & 3.5481 \\
\midrule
Lack of Panx1 increases adipocyte hypertrophy and reduces adipocyte numbers in subcutaneous fat in vivo. & With both a normal diet, and a a high-fat diet, a lack of Panx1 increases cell size. & 3.5618 \\

\bottomrule
\end{tabular}
}
\caption{Borderline IMS Model prediction samples. We note that 3 appears to be a good threshold for matching, as pairs with an IMS over 3 tend to discuss the same scientific findings.}
\label{tab:prediction_samples}
\end{table*}

\section{Regression details}
\label{sec:regression_details}

\begin{table*}
\begin{center}
\begin{tabular}{llll}
\hline
Model:            & MixedLM & Dependent Variable: & paper\_sentence\_score  \\
No. Observations: & 1111150 & Method:             & REML                    \\
No. Groups:       & 6705    & Scale:              & 0.1084                  \\
Min. group size:  & 31      & Log-Likelihood:     & -349944.7797            \\
Max. group size:  & 67063   & Converged:          & Yes                     \\
Mean group size:  & 165.7   &                     &                         \\
\hline
\end{tabular}
\end{center}

\begin{center}
\begin{tabular}{lrrrrrr}
\hline
                         & $\beta$ Coef. & Std.Err. &       z & P$> |$z$|$ & [0.025 & 0.975]  \\
\hline
\emph{Intercept}                &  3.299 &    0.007 & 489.729 &       0.000 &  3.286 &  3.312  \\
Outlet Type: Press Release &  0.037 &    0.001 &  31.187 &       0.000 &  0.035 &  0.039  \\
Outlet Type: Science \& Technology      &  0.034 &    0.001 &  30.581 &       0.000 &  0.032 &  0.036  \\
Field: Biology                  & -0.018 &    0.020 &  -0.904 &       0.366 & -0.056 &  0.021  \\
Field: Psychology               &  0.040 &    0.018 &   2.168 &       0.030 &  0.004 &  0.076  \\
Field: Medicine                 &  0.206 &    0.017 &  11.813 &       0.000 &  0.171 &  0.240  \\
Field: Computer\_science        &  0.050 &    0.024 &   2.132 &       0.033 &  0.004 &  0.096  \\
Group Var                &  0.009 &    0.001 &         &             &        &         \\
\hline
\end{tabular}
\end{center}
    \caption{Regression table for RQ1} 
    \label{tab:RQ1_regression}
\end{table*}

\begin{table*}
\begin{center}
\begin{tabular}{llll}
\hline
Model:            & MixedLM & Dependent Variable: & paper\_sentence\_score  \\
No. Observations: & 182735  & Method:             & REML                    \\
No. Groups:       & 1360    & Scale:              & 0.1525                  \\
Min. group size:  & 31      & Log-Likelihood:     & -89654.8514             \\
Max. group size:  & 89523   & Converged:          & Yes                     \\
Mean group size:  & 134.4   &                     &                         \\
\hline
\end{tabular}
\end{center}

\begin{center}
\begin{tabular}{lrrrrrr}
\hline
                            & $\beta$ Coef. & Std.Err. &       z & P$> |$z$|$ & [0.025 & 0.975]  \\
\hline
\textit{Intercept}                   &  3.777 &    0.013 & 292.571 &       0.000 &  3.752 &  3.803  \\
Is Verified  User?  & -0.047 &    0.004 & -11.044 &       0.000 & -0.056 & -0.039  \\
Is Organizational Account? & 0.042 &    0.002 & -19.026 &       0.000 & -0.046 & -0.037  \\
User Metric: log(Followers)             & -0.003 &    0.001 &  -5.059 &       0.000 & -0.004 & -0.002  \\
User Metric: log(Following)          &  0.000 &    0.001 &   0.369 &       0.712 & -0.001 &  0.002  \\
User Metric: Account Age (in years)   &  0.004 &    0.000 &  10.824 &       0.000 &  0.003 &  0.005  \\
Field: Biology                     & -0.025 &    0.030 &  -0.850 &       0.395 & -0.083 &  0.033  \\
Field: Psychology                  &  0.308 &    0.028 &  11.052 &       0.000 &  0.254 &  0.363  \\
Field: Medicine                    &  0.206 &    0.026 &   7.826 &       0.000 &  0.155 &  0.258  \\
Field: Computer\_science           & -0.352 &    0.035 & -10.158 &       0.000 & -0.420 & -0.284  \\
Group Var                   &  0.059 &    0.006 &         &             &        &         \\
\hline
\end{tabular}
\end{center}
    \caption{Regression table for RQ2} 
    \label{tab:RQ2_regression}
\end{table*}

\autoref{tab:RQ1_regression} shows the regression table for RQ1.
\autoref{tab:RQ2_regression} shows the regression table for RQ2.

\end{document}